\documentclass[letterpaper]{article} 
\usepackage{aaai25}
\usepackage{times}  
\usepackage{helvet}  
\usepackage{courier}  
\usepackage[hyphens]{url}  
\usepackage{graphicx} 
\usepackage{natbib}  
\usepackage{caption} 
\usepackage{algorithm}
\usepackage{algorithmic}
\usepackage{newfloat}
\usepackage{listings}

\DeclareCaptionStyle{ruled}{labelfont=normalfont,labelsep=colon,strut=off} 
\floatstyle{ruled}
\newfloat{listing}{tb}{lst}{}
\floatname{listing}{Listing}

\usepackage{xtab}
\extrafloats{300}

\usepackage{bbding}
\usepackage{booktabs}
\usepackage{array}
\usepackage{makecell}
\usepackage{amsmath}

\usepackage{latexsym}
\usepackage{tabularx}
\usepackage{graphicx}
\usepackage{booktabs}
\usepackage{multirow}
\usepackage{multicol}
\usepackage{fontawesome5}
\usepackage{booktabs}
\usepackage{multirow}
\usepackage[table,xcdraw]{xcolor}

\usepackage{longtable} 
\usepackage{array} 
\usepackage[T1]{fontenc}

\usepackage[utf8]{inputenc}
\usepackage{xcolor}
\definecolor{darkgreen}{rgb}{0.0, 0.7, 0.0}
\usepackage{microtype}

\usepackage{inconsolata}
\usepackage{booktabs} 

\usepackage{xcolor}

\usepackage{xcolor}

\title{Fluent but Unfeeling: The Emotional Blind Spots of Language Models}

\author{
\textbf{Bangzhao Shu}$^{1}$\thanks{Equal contribution},
\textbf{Isha Joshi}$^{1}$\footnotemark[1],
\textbf{Melissa Karnaze}$^{2}$,
\textbf{Anh C. Pham}$^{3}$,
\textbf{Ishita Kakkar}$^{3}$,
\textbf{Sindhu Kothe}$^{2}$,
\textbf{Arpine Hovasapian}$^{4}$,
\textbf{Mai ElSherief}$^{1}$\\
\normalfont
$^{1}$Northeastern University\\
$^{2}$UC San Diego\\
$^{3}$University of Massachusetts Amherst\\
$^{4}$Independent Researcher\\
{\{shu.b, joshi.ishaa, m.elsherif\}@northeastern.edu},
{\{mkarnaze, skothe\}@ucsd.edu},\\
{\{acpham, ikakkar\}@umass.edu}
}

\begin{document}
\maketitle
\begin{abstract}

The versatility of Large Language Models (LLMs) in natural language understanding has made them increasingly popular in mental health research. While many studies explore LLMs' capabilities in emotion recognition, a critical gap remains in evaluating whether LLMs align with human emotions at a fine-grained level. Existing research typically focuses on classifying emotions into predefined, limited categories, overlooking more nuanced expressions. To address this gap, we introduce EXPRESS, a benchmark dataset curated from Reddit communities featuring 251 fine-grained, self-disclosed emotion labels. Our comprehensive evaluation framework examines predicted emotion terms and decomposes them into eight basic emotions using established emotion theories, enabling a fine-grained comparison. Systematic testing of prevalent LLMs under various prompt settings reveals that accurately predicting emotions that align with human self-disclosed emotions remains challenging. Qualitative analysis further shows that while certain LLMs generate emotion terms consistent with established emotion theories and definitions, they sometimes fail to capture contextual cues as effectively as human self-disclosures. These findings highlight the limitations of LLMs in fine-grained emotion alignment and offer insights for future research aimed at enhancing their contextual understanding.

\end{abstract}
\begin{figure}[!htb]
    \centering
    \includegraphics[width=0.4\textwidth, height=7.5cm]{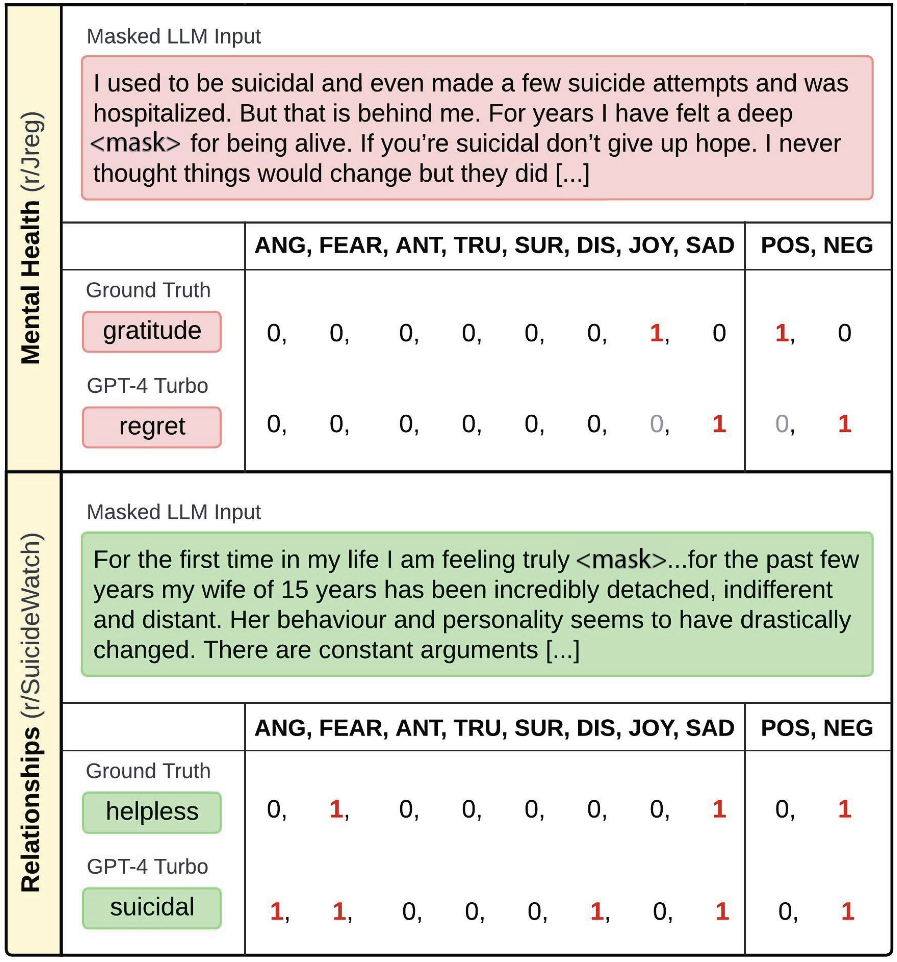}
    \caption{An example of EXPRESS and our emotion recognition evaluation framework is presented. Language models are prompted to predict masked emotions in the text. Both the predicted and self-disclosed emotion words are then decomposed into eight basic emotion dimensions and two sentiment dimensions. The differences between the predictions and the self-disclosed emotions at the dimensional level indicate that GPT-4 Turbo fails to accurately capture the emotions conveyed in human self-disclosures.}
    \label{fig:bias_examples}
\end{figure}

\section{Introduction}
LLMs have been trained on vast amounts of written human language (e.g., from the internet), some of which contains descriptions of emotional experiences and emotional discourse~\cite{achiam2023gpt,brown2020language}. LLMs have been designed to interface with users with some knowledge of human emotion~\cite{wang2023emotional}. From an evolutionary standpoint, universal human emotions, like fear, joy, and disgust, developed to solve unique sets of challenges faced by ancestral humans, and LLMs can learn about emotions through training.

Specifically, LLMs can be assessed on competencies in \textit{emotional intelligence}, a construct that encompasses how people can respond strategically to and leverage their emotional experiences in productive ways, regardless of how adaptive or ill-suited to a modern-day situation an emotional reaction might initially be~\cite{brackett2011emotional,salovey1990emotional}. While LLMs cannot possess complete emotional intelligence, their stochastic outputs can be evaluated in two of the four domains of emotional intelligence as posited by~\citeauthor{salovey1990emotional}: accurately perceiving what emotion is being expressed by a human via their written expressions of emotional disclosures; and demonstrating accurate analysis of what a human is likely experiencing given contextual cues (such as a person’s situation, appraisals of that situation, and/or corresponding bodily responses). The extent to which frontier foundation models can perform such tasks is an empirical question. As shown in Figure~\ref{fig:bias_examples}, the LLM's predicted emotions fail to align with self-disclosed emotions when asked to provide completions to various human experience prompts.

Accurate emotion recognition has the potential to substantially enhance a broad spectrum of natural language processing (NLP) tasks. By integrating fine-grained affective understanding, dialogue systems can become more emotionally aware, thereby improving their capacity to comprehend and generate human-like emotions~\cite{liu-etal-2021-towards}. Moreover, emotion recognition facilitates quantitative analyses of social dynamics, such as political discourse, customer service, and public opinion mining. Finally, it also enables critical NLP-driven applications, including automated depression screening~\cite{dan2024}.

Substantial research efforts have been made to study the emotion recognition capabilities of LLMs. These studies usually focus on constructing various emotion recognition benchmarks from different data sources, domains, and emotion theories~\cite{strapparava2007semeval, chatterjee2019semeval}. However, current benchmarks face three main \textbf{limitations}. First, little attention has been given to the self-disclosure of emotional experiences, with existing research typically relying on crowdsourcing or expert annotations to label emotions in text, which might lead to unreliable evaluations~\cite{singh2023language, sabour2024emobench}. Second, these studies are often limited by a short, predefined list of emotions, restricting their ability to capture fine-grained nuances in emotional experiences~\cite{feng2021emowoz}. Third, current benchmarks primarily focus on short sentences, contexts, or dialogues, neglecting the self-disclosure of emotions in longer contexts~\cite{demszky2020goemotions}.

To address this gap, we construct a new benchmark, \includegraphics[width=1.2em, height=1.2em]{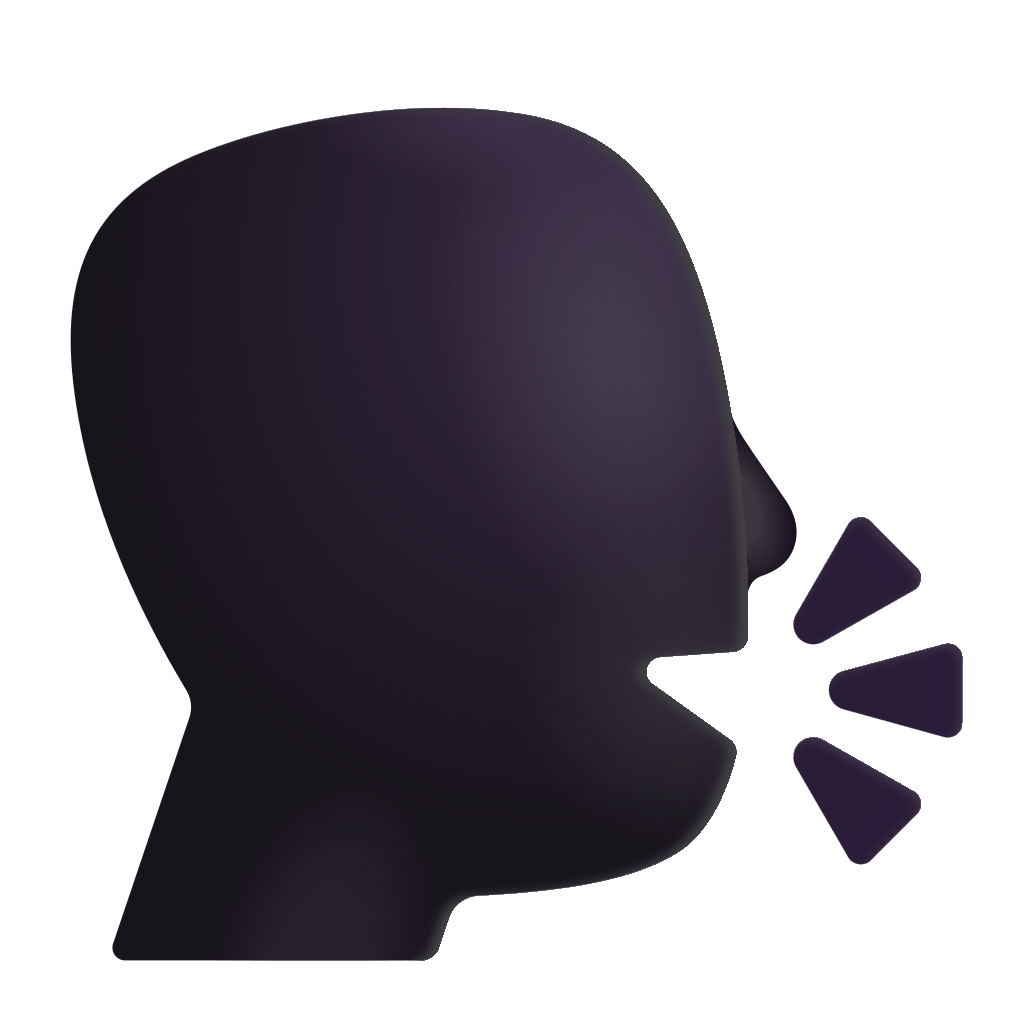} EXPRESS (\textbf{EX}periences and \textbf{PR}ocessed \textbf{E}motions in \textbf{S}elf-disclosure \textbf{S}tories), which consists of 33,679 human experiences and their associated self-disclosed emotions. The assessment framework is illustrated in Figure~\ref{fig:bias_examples}. In this framework, we mask emotion words in human self-disclosure texts and prompt the LMs to predict the masked words. To move beyond predefined emotion classes and evaluate models' emotion recognition capabilities at a finer granularity, we decompose the emotions into 10 dimensions of basic emotions and sentiments based on Plutchik’s Wheel~\cite{plutchik1980general, mohammad2013crowdsourcing}. Using EXPRESS, we benchmark the emotion recognition performance of 14 language models.

Our results reveal that the emotion recognition capabilities of LMs correlate with model size, model family, model architecture, and prompting strategies. While the accuracy of predicting emotions at the lexical level is low, some LMs demonstrate the ability to capture the basic emotions underlying compound emotions, even when the predicted emotion word does not exactly match the label. Furthermore, we find that LLMs are capable of in-context learning for emotion recognition when provided with examples. We also find that, under the best settings, LLMs are able to generate emotions consistent with theoretical definitions, but they are sometimes less contextually ``aware'' than the emotions self-disclosed by humans. These findings are particularly valuable for the development of emotion-aware AI systems, especially in future mental health applications~\cite{hua2024large, ji2022mentalbert, adhikary2024exploring}.

This paper makes the following contributions:
\begin{itemize} 
\item We present EXPRESS, a novel benchmark designed for emotion recognition that serves as a resource for evaluating models' emotion recognition capabilities and facilitating potential emotion alignment. 
\item We propose a multi-faceted emotion recognition evaluation framework that encompasses multiple metrics: lexical accuracy, decomposed emotion vector accuracy, and contextual accuracy. 
\item Our framework incorporates various prompting strategies, including few-shot learning and Chain of Thought prompting, to evaluate models' emotional reasoning and in-context learning capabilities. We release our code and dataset on GitHub: \url{https://github.com/Computing-for-Social-Good-CSG/express-emotion-recognition.git}
\end{itemize}

\begin{table*}[!ht]
\centering
{\fontsize{7}{8}\selectfont
\resizebox{\textwidth}{!}{%
    \begin{tabular}{>{\raggedright\arraybackslash}p{35mm} >{\raggedright\arraybackslash}p{20mm} >{\raggedright\arraybackslash}p{35mm} r p{10mm} p{18mm} >{\raggedright\arraybackslash}p{15mm} >{\raggedright\arraybackslash}p{15mm} }
    \toprule
    \textbf{Work/Dataset} & \textbf{Source} & \textbf{Domain} & \textbf{Size} & \textbf{No. Emotions} & \textbf{Avg Context Length} & \textbf{Topic Diversity} & \textbf{Annotator} \\ \midrule

    GoEmotions~\cite{demszky2020goemotions} & Reddit & General Reddit Comments & \textbf{58,009} & 27 & 13 & Unknown & Crowdsourced \\ \arrayrulecolor{black!20}\hline

    EmoTrigger~\cite{singh2023language} & CancerNet, Twitter, Reddit & Health-related, Natural disasters, General Reddit posts & 900 & 8 & 22 & 5 & Expert \\ \hline

    EmoWOZ~\cite{feng2021emowoz} & Amazon MTurk & Human-human + human-machine conversations & 11,000 & 7 & 187 per dialogue 13 per utterance & 7 & Expert \\ \hline

    SemEval Task \#14~\cite{strapparava2007semeval} & NYT, CNN, BBC, Google News & News Headlines & 1,250 & 6 & 7 & 13 & Trained \\ \hline

    EmoContext~\cite{chatterjee2019semeval} & Twitter & Conversational agent interactions & 38,424 & 4 & 23 & Unknown & Trained \\ \hline

    ISEAR~\cite{scherer1994evidence} & Survey & Personal narratives on major emotional events & 7,665 & 7 & 23 & 31 & \textbf{Self-reported} \\ \hline

    DailyDialogue~\cite{li2017dailydialog} & Various Websites & Everyday multi-turn dialogues & 13,118 & 6 & 115 per dialogue 15 per utterance & 10 & Expert \\ \hline

    EmoBank~\cite{buechel2022emobank} & MASC + SemEval-2007 & News headlines, blogs, fiction, etc. & 10,548 & 9 & 15 & 33 & Crowdsourced \\ \hline

    CrowdFlower~\cite{van2012designing} & Twitter & General Twitter tweets & 39,740 & 13 & 36 & 9 & Crowdsourced \\ \hline

    EmoInt~\cite{mohammad2017emotion} & Twitter & General Twitter tweets & 7097 & 4 & 16 & 46 & Crowdsourced \\ \hline

    DENS~\cite{liu2019dens} & Gutenberg, Wattpad & Long-form English narratives & 9,710 & 8 & 86 & 7 & Crowdsourced + Expert \\ \hline

    Emotion-Stimulus~\cite{ghazi2015detecting} & FrameNet & Blogs & 820 & 9 & 18 & 28 & Expert \\ \hline

    Tales Emotion~\cite{Alm2005PerceptionsOE} & Grimm Fairy Tales & Blogs & 15,302 & 8 & 21 & 10 & Trained \\ \hline

    XED~\cite{ohman2018creating} & OPUS & Movie subtitles & 33,548 & 8 & 9 & Unknown & Crowdsourced \\ \arrayrulecolor{black}\midrule

    \textbf{EXPRESS (ours)} & Reddit & Reddit Posts & 33,679 & \textbf{251} & \textbf{259} & \textbf{49} & \textbf{Self-disclosed} \\

    \bottomrule
    \end{tabular}
}}
\caption{Comparison of emotion datasets in terms of size, domain, number of emotions, average text length, topic diversity, and annotation method.}
\label{tab:lit_review}
\end{table*}

\section{Related Work}
\subsection{Emotion Taxonomy}

While emotion theorists have nuanced definitions of emotions that may differ in their descriptions of emotional features, most agree that emotions are functional in preparing us to respond to perceived or real environmental stimuli~\cite{gross2011emotion}. Theories of emotion define discrete emotional experiences or reactions to events (real or imagined) as representations that may include correlated physiological responses, appraisal processes, subjective feelings, and action tendencies~\cite{schiller2024human}.

Many theories of emotion are studied in the field of affective science, but computational models of emotion in NLP have primarily been based on two families of theories. The first views emotions as fixed atomic units, often referred to as basic emotions, while the second conceptualizes emotion as existing within a multidimensional space~\cite{dan2024}. Within the family of fixed atomic unit theories, one prominent theory proposes six basic emotions: surprise, happiness, anger, fear, disgust, and sadness~\cite{ekman1999basic}. Another theory, known as the Plutchik Wheel of Emotion, posits that emotions consist of eight basic emotions in four opposing pairs: joy-sadness, anger-fear, trust-disgust, and anticipation-surprise~\cite{plutchik1980general}.

Recent advances in psychology have introduced new conceptual and methodological approaches to capturing the more complex “semantic space” of emotion, which aligns with the second family of theories~\cite{cowen2019}. These models typically use two dimensions—valence and arousal—to describe emotions~\cite{russell1980circumplex}. Some models further include a third dimension, dominance, to provide additional nuance in describing emotions~\cite{fontaine2007nottwo}.

\subsection{Emotion Recognition Benchmarks}

Substantial datasets for emotion recognition exist, each with varying emotion label domains. Table~\ref{tab:lit_review} summarizes key characteristics of existing datasets, including size, number of emotion labels, average context length, and annotation methods.

\noindent\textbf{Emotion label coverage is often limited.} Most datasets include only a small set of predefined emotion categories. For example, Emotion-Stimulus~\cite{ghazi2015detecting}, Tales Emotions~\cite{Alm2005PerceptionsOE}, and SemEval Task 14~\cite{strapparava2007semeval} rely on Ekman’s six basic emotions~\cite{ekman1999basic}, sometimes supplemented with a few additional labels. XED~\cite{ohman2018creating} and EmoTrigger~\cite{singh2023language} use Plutchik’s eight primary emotions. These limited sets of emotion labels, focused on a coarse level of classification, restrict the ability to study fine-grained emotional nuances.

\noindent\textbf{Context lengths tend to be short.} Many datasets are built on platforms such as Twitter or Reddit comments, or focus on dialogue utterances and news headlines. As a result, the average context length for emotion recognition typically ranges from 10 to 36 words, with CrowdFlower reaching the upper bound at 36 words~\cite{van2012designing}.

\noindent\textbf{Dataset sizes vary.} While small datasets like Emotion-Stimulus contain under 1,000 examples, larger resources such as EmoContext~\cite{chatterjee2019semeval}, DailyDialogue~\cite{li2017dailydialog}, and GoEmotions~\cite{demszky2020goemotions} include tens of thousands of samples.

\noindent\textbf{Annotations are typically crowdsourced or expert-labeled.} Many benchmarks rely on either expert labeling or crowdsourcing, where annotators infer emotions from external text. ISEAR~\cite{scherer1994evidence} is an exception that uses self-reported emotion events, but its ecological validity is limited by its collection method: participants were asked to describe experiences for a fixed list of seven predefined emotions, constraining natural emotional expression and overlooking subtle or more complex feelings. In addition, relying on external annotations limits the granularity of emotion labels.

Prior research provides a foundation for assessing the emotion recognition capabilities of LLMs, though it faces the limitations mentioned above. Our work addresses these issues by scaling up context length to an average of 259 words and the range of emotion labels to 251. We use self-disclosed emotions as ground truth labels without external annotation because they are considered ecologically valid. As naturally occurring disclosures, they allow individuals to freely and authentically share their internal experiences, including emotional reactions to past events, without being constrained by predefined categories or researcher-led methods~\cite{pennebaker1986confronting,frattaroli2006experimental,davitz2013language}. Moreover, self-report remains a cornerstone of methods for empirically investigating subjectively felt emotional experiences~\cite{mauss2009measures}, further supporting the use of self-disclosed emotions as ground truth. Additionally, our framework allows models to generate context-based, non-predefined emotions, ensuring sufficient variation in emotional nuances. Together, these improvements establish our dataset as an ecologically valid and fine-grained benchmark for evaluating emotion recognition capabilities in language models.

\section{EXPRESS: A Comprehensive Benchmark for Emotion Recognition}

\begin{table*}[!htb]
\centering
  \scriptsize
   \begin{tabular}{@{}p{0.8\textwidth}|p{0.15\textwidth}}
     \toprule
    \textbf{Experience Contextualized Prompts} & \textbf{Self-disclosed Emotion} \\
    \midrule
    Feeling <mask> after getting a role in a movie. [...], now am scared cause the director doesn't know that I got no acting experiences or skills (am very bad at memorising my lines) if he finds out he won't hire me for the movie. I don't know what to do. & afraid \\

    \hline
    How do I do thiS... I feel <mask>. Battling so many health issues right now, mostly gi related. [...] I am incredibly incredibly <mask> and going through 1 to 5 juul pods a day... I feel like there’s no way out of this. Any advice? I can’t imagine my depression and fatigue getting worse...  & panicked, depressed\\
 \bottomrule
  \end{tabular}
  \caption{Examples of naturally occurring emotionally-centered prompts in EXPRESS. Self-disclosed emotions (ground truth) are replaced by <mask> token.}
  \label{tab:dataset_examples}
  
\end{table*}

\textbf{Selection of an Emotion Lexicon.}
The Berkeley Well-Being Institute synthesized a complete list of 271 emotions~\cite{berkeleywellbeingListEmotions} based on multiple emotion theories: Discrete, Circumplex~\cite{russell1980circumplex}, Plutchik's Wheel~\cite{plutchik1980general}, and other emotion theories. Instead of selecting a single emotion theory, we used the Berkeley Well-Being list of emotions because it combines multiple theories and represents the largest available emotion lexicon. A primer on the emotion theories used in this paper is provided in the Appendix.

\noindent\textbf{Collecting self-disclosed experiences and emotions.} Our primary objective is to assess whether LMs can predict emotions based on real-life nuanced experiences. To achieve this, we created prompts that embed a self-disclosed emotion. To ensure we evaluate LMs in scenarios that mirror actual language usage, we construct our prompts from natural contexts that we retrieve from Reddit, rather than crowdsourcing prompts. 

Because of its pseudonymity, Reddit is a popular platform for discussing real-life experiences. Reddit served as the primary data source due to its support for longer posts (up to 40,000 characters) which enabled the collection of rich and nuanced human experiences and the corresponding evoked emotions. The Reddit API Praw~\cite{redditPRAW} \footnote{BSD 2-Clause License: licensed under a permissive license allowing redistribution and modification with the retention of copyright and disclaimer notices} was utilized to collect posts from all subreddits containing at least one emotion from the Berkeley Well-Being list.

\noindent\textbf{Emotion Masking.} To mask the self-disclosed emotions in the collected posts, we designed a comprehensive regular expression protocol, as not all emotion keywords in a post are self-disclosed by the author. For example, the author might use emotion keywords to describe external events or other people’s feelings rather than their own. Our protocol includes three main patterns: `I feel + emotion', `I am + emotion', and `no-pronoun + feeling + emotion'. To make the algorithm robust to variations in natural language phrasing, we designed a series of rules, with details provided in Appendix.

We included the pattern `feel' because prior work indicates that humans use the word `feeling' interchangeably with `emotion', even though feelings and emotions are not the same~\cite{berkeleywellbeingListEmotions}. Feelings encompass both emotional experiences (e.g., feeling sad) and physical sensations (e.g., feeling hungry). This distinction justifies our use of the following pattern-matching formats: (\textit{I + feel/am + emotion}) and (\textit{no-pronoun + feeling + emotion}).

Due to the context window size limitations of some language models and the fact that some posts are extremely lengthy, we segmented the posts into chunks of 512 tokens. During the segmentation process, we ensured that the context surrounding the target masked emotions was maximized. If multiple masked emotions existed, they were grouped based on their relative positions in the text. Table~\ref{tab:algo} in Appendix outlines the algorithm used to perform post segmentation.

The resulting dataset (EXPRESS) comprises 33,697 posts with an average word count of 259 per post. EXPRESS posts originate from 6,930 unique subreddits and span the time period from June 2009 - April 2024. More details of the dataset are provided in the Appendix (Table~\ref{table:dataset_stats}, Table~\ref{tab:dataset_examples_appendix}, Table~\ref{tab:top60_emotions}). Across the dataset, a total of 52,632 emotion words were identified covering 251 (92.62\%) out of 271 Berkeley Well-Being emotions. We create prompts from EXPRESS by replacing the original emotion with a <mask> token. Table~\ref{tab:dataset_examples} depicts examples from our dataset.
\section{Evaluating Emotion Recognition Capabilities of LLMs}

Using EXPRESS, we measured the emotion recognition capabilities of 14 prevalent language models including four masked language models, three Seq2Seq language models, and seven causal language models.

\subsection{Model Details}

Using our dataset, we evaluated several variants of open-source and closed-source language models widely used in current research. We included four prevalent masked language models, as they are specifically designed for masked language modeling tasks. These models are RoBERTa-base~\cite{liu2019robertarobustlyoptimizedbert}, Longformer~\cite{beltagy2020longformerlongdocumenttransformer}, Mental-RoBERTa~\cite{ji2022mentalbert}, and Mental-Longformer~\cite{ji2023mentallongformer}, the latter two of which have been further pre-trained on mental health-related corpora. For Seq2Seq language models, we included three models from the Flan-T5 family (large, XL, and XXL)~\cite{chung2022flant5}. For causal language models, we focused on instruction-tuned models of varying sizes, as they are fine-tuned to follow instructions. These include Llama3.1-Instruct (8B, 70B)~\cite{grattafiori2024llama3herdmodels}, Gemma2-Instruct (2B, 9B, 27B)~\cite{gemmateam2024gemma2improvingopen}, GPT-3.5-turbo, and GPT-4o~\cite{openai2024gpt4technicalreport}. The temperature was set to 0.0 for all experiments to minimize the effects of randomness.

\subsection{Experimental Setup}

We evaluated the performance of the four masked language models by directly filling in the masked emotions. For the remaining 10 models, we prompted them to predict the masked emotions based on the context. We designed the experiments with four different settings: zero-shot, few-shot with 4 random examples, few-shot with 4 nearest examples, and Chain-of-Thought (CoT) prompting~\cite{wei2022chain}.

The zero-shot setting served as the basic test of the emotion recognition ability of LLMs based on self-disclosed emotional experiences. Models were directly instructed to predict the <mask> word with an emotion based on the context. The prompt template used in this setting is detailed in Appendix (Table~\ref{tab:combined-prompt}).

To investigate whether LLMs can enhance their emotion detection ability by learning from examples, we included two few-shot settings. In both settings, we used four examples, as prior work has shown that using a larger number of exemplars does not significantly improve model performance~\cite{min2022rethinking}. Additionally, we ensured that the number of <mask> tokens in the exemplars matched that of the target post to avoid confounding effects. The first few-shot setting used four examples randomly selected from EXPRESS, while the second used the Bert-base-uncased model~\cite{devlin2019bert} to compute sentence embeddings and applied Euclidean distance to find the four nearest examples to the Reddit posts~\cite{liu2021makes}. By comparing these two settings, we aimed to explore whether providing similar experiences in the examples could further enhance the models' ability.

Studies have shown that CoT prompting improves performance across a range of arithmetic, commonsense, and symbolic reasoning tasks~\cite{wei2022chain}. However, some studies have also suggested that CoT prompting may not enhance performance in socially sensitive domains, such as addressing harmful questions~\cite{shaikh-etal-2023-second}. In this work, we included the CoT prompting setting to examine whether it could further improve the models' ability to predict emotions.

\subsection{Measuring Accuracy of Emotion Recognition}

To evaluate LMs’ emotion recognition capabilities on a fundamental level, beyond calculating lexicon accuracy, we adopted the approach of the NRC Emotion Lexicon (EmoLex)~\cite{mohammad2013crowdsourcing}. EmoLex is a widely used resource that analyzes 14,182 unigrams and associates these unigrams, through crowdsourcing, with eight basic emotions—anger, anticipation, disgust, fear, joy, sadness, surprise, and trust—as well as with positive and negative sentiment. The associations are represented as binary scores (0 or 1), indicating whether a word is linked to a particular emotion or sentiment. We leveraged EmoLex to construct vector representations for each predicted and actual emotion. These vectors, as shown in Figure \ref{fig:bias_examples}, are 10-dimensional: 8 dimensions correspond to the basic emotions, and 2 represent positive and negative sentiment. By converting all words into these 10-dimensional vectors, we evaluated the model-predicted emotions against the self-disclosed emotions on a basic emotion and sentiment level.

During this process, some model-generated emotions were not included in the NRC Emotion Lexicon. To address this, we replicated EmoLex's crowdsourcing task on Amazon Mechanical Turk (AMT) to generate vector representations for these additional emotions. Further details are provided in Appendix (Figure~\ref{fig:turk-one}).

We evaluated the results using three metrics: (1) Lexical Accuracy ($Acc_L$), defined as the percentage of exact lexical matches ($N_{lm}$) from all masks ($N$): $Acc_L = N_{lm}/N$; (2) Vector Accuracy ($Acc_V$), defined as the percentage of exact vector matches ($N_{vm}$) between the 10-dimensional basic emotion vectors for the predicted and actual emotions across all masks: $Acc_V = N_{vm}/N$; and (3) Average Vector F-1 score ($F1_V$), which balances precision and recall to evaluate the model's ability to predict each dimension of the 10-dimensional emotion vector. More details on the F1-score calculation for vectors are provided in the Appendix section titled Evaluation Metrics.

We included $Acc_V$ as a metric because predicted and actual emotions may not align lexically due to the diversity of emotion vocabulary but could still match at the basic emotion level. For example, `angry' and `furious' share the same emotion vector but are two different labels lexically. A higher $Acc_V$ indicated greater alignment between the actual and predicted emotions. Similarly, $F1_V$ was included to assess how closely the predicted emotions approximated the self-disclosed ones, even when there was no exact match across all dimensions.

\section{Results}

\begin{table*}[t]
\setlength\tabcolsep{3pt}
\setlength\tabcolsep{5pt}
\fontsize{8}{7}\selectfont
\centering
\caption{Performance Comparison of Language Models Across Different Evaluation Settings. Metrics include Lexical Accuracy ($Acc_L$), Vector Accuracy ($Acc_V$), and Average Vector F-1 Scores ($F1_V$) across Zero-shot, Few-shot (random and nearest examples), and CoT settings.}
\label{tab:results}
\begin{tabularx}{\textwidth}{@{}l c c l ccc ccc ccc ccc@{}}
\toprule
\multirow{2}{*}{\makecell{\textbf{Model}\\\textbf{Architecture}}} & \multicolumn{2}{c}{\multirow{2}{*}{\textbf{Model}}} & \multicolumn{3}{c}{\textbf{Zero-shot}} & \multicolumn{3}{c}{\textbf{Few-shot (random)}} & \multicolumn{3}{c}{\textbf{Few-shot (nearest)}} & \multicolumn{3}{c}{\textbf{CoT}} \\
\cmidrule(r){4-6} \cmidrule(r){7-9} \cmidrule(r){10-12} \cmidrule(r){13-15}
& \multicolumn{2}{c}{} & \textbf{$Acc_L$} & \textbf{$Acc_V$} & \textbf{$F1_V$} & \textbf{$Acc_L$} & \textbf{$Acc_V$} & \textbf{$F1_V$} & \textbf{$Acc_L$} & \textbf{$Acc_V$} & \textbf{$F1_V$} & \textbf{$Acc_L$} & \textbf{$Acc_V$} & \textbf{$F1_V$} \\
\midrule
\multirow{4}{*}{\textbf{Masked LMs}} & \multirow{2}{*}{RoBERTa} & base & \textbf{0.318} & 0.369 & 0.658 &{-} & {-}& {-}&{-} &{-} &{-} &{-} &{-} &{-} \\
& & mental & 0.313 & 0.366 & 0.654 &{-} & {-}& {-}&{-} &{-} &{-} &{-} &{-} &{-} \\

\addlinespace[4pt]

& \multirow{2}{*}{Longformer} & base & 0.309 & 0.358 & 0.645 &{-} & {-}& {-}&{-} &{-} &{-} &{-} &{-} &{-} \\
& & mental & 0.277 & 0.330 & 0.633 &{-} & {-}& {-}&{-} &{-} &{-} &{-} &{-} &{-} \\
\midrule
\multirow{3}{*}{\textbf{Seq2Seq}} & \multirow{3}{*}{Flan-T5} & large & 0.051 & 0.097 & 0.434 &{-} & {-}& {-}&{-} &{-} &{-} &{-} &{-} &{-} \\
& & xl & 0.063 & 0.111 & 0.471 &{-} & {-}& {-}&{-} &{-} &{-} &{-} &{-} &{-} \\
& & xxl & 0.102 & 0.167 & 0.532 &{-} & {-}& {-}&{-} &{-} &{-} &{-} &{-} &{-} \\
\midrule
\multirow{8}{*}{\textbf{Causal LMs}} & \multirow{2}{*}{Llama-3.1} & 8B & 0.149 & 0.222 & 0.591 & 0.175 & 0.248 & 0.619 & 0.195 & 0.265 & 0.624 & 0.153 & 0.228 & 0.586\\
& & 70B & 0.264 & 0.338 & 0.675 & 0.265 & 0.343 & 0.677 & 0.279 & 0.356 & 0.683 & 0.198 & 0.272 & 0.650\\

\addlinespace[4pt]

& \multirow{3}{*}{Gemma-2} & 2B & 0.078 & 0.147 & 0.530 & 0.105 & 0.174 & 0.550 & 0.122 & 0.193 & 0.557 & 0.052 & 0.110 & 0.514\\
& & 9B & 0.223 & 0.293 & 0.639 & 0.268 & 0.338 & 0.673 & 0.278 & 0.347 & 0.677 & 0.152 & 0.200 & 0.634\\
& & 27B & 0.268 & 0.341 & 0.667 & 0.290 & 0.358 & 0.675 & 0.298 & 0.366 & 0.678 & 0.154 & 0.211 & 0.636\\

\addlinespace[4pt]

& GPT & 3.5-turbo & 0.217 & 0.285 & 0.629 & 0.247 & 0.313 & 0.647 & 0.250 & 0.315 & 0.647 & 0.176 & 0.244 & 0.602\\
& & 4o & 0.313 & \textbf{0.388} & \textbf{0.711} & 0.350 & 0.422 & 0.732 & \textbf{0.364} & \textbf{0.436} & \textbf{0.738} & 0.310 & 0.383 & 0.723\\

\addlinespace[4pt]

& Average (Causal) & & 0.210 & 0.272 & 0.647 & 0.242 & 0.314 & 0.653 & 0.255 & 0.325 & 0.658 & 0.171 & 0.235 & 0.621\\
\midrule

\multirow{2}{*}{\textbf{Baselines}} & Random & & {-} & 0.001 & 0.322 & {-}& {-}& {-}& {-}& {-}& {-}& {-}& {-}&{-}\\
& All-zeros & & {-}& 0.037 & 0.000 & {-}& {-}& {-}& {-}& {-}& {-}& {-}& {-}&{-}\\

\bottomrule
\end{tabularx}
\end{table*}

Here, we present our findings on the emotion recognition capabilities of LLMs evaluated on the EXPRESS dataset. \\
\\
\noindent\textbf{Fine-Grained Emotion Alignment is Challenging for LMs.} Models demonstrated significant variability in their ability to predict emotions from emotional experiences in the zero-shot setting, as shown in Table~\ref{tab:results}. $Acc_L$ ranged from 0.051 to 0.318, while $Acc_V$, slightly higher, ranged from 0.097 to 0.388. $F1_V$ ranged from 0.434 to 0.711, compared to a baseline of randomly generated vectors at 0.322. Overall, the results show that it is challenging to predict human self-disclosed emotions from emotional experiences. The substantial number of emotion words, along with the similarity and overlapping nature of some emotion terms, may contribute to the low $Acc_L$. However, $Acc_V$, which evaluates the decomposed vectors of emotion terms, remains relatively low, increasing by only about 0.06 on average. This indicates that for most models, language models fail to align with human self-disclosed emotions on at least one basic emotion or sentiment dimension in the majority of predictions.

Some models, such as Flan-T5-large, XL, and Gemma2-2B, have $Acc_V$ around 0.1, meaning they can only accurately predict around 10\% of the emotions. Their $F1_V$ ranges from 0.4 to 0.5, which, although higher than randomly generated vectors, indicates their limited ability to correctly predict emotions. On the other hand, some models demonstrate relatively better emotion recognition ability, including the four masked language models, Llama-3.1-70B, Gemma-2-27B, and GPT-4o. These models achieve $Acc_V$ over 0.3 and $F1_V$ exceeding 0.6, indicating a certain degree of alignment with self-disclosed emotions or, at the very least, some aspects of them.

\begin{figure*}[t]
    \centering
    \includegraphics[width=0.9\textwidth]{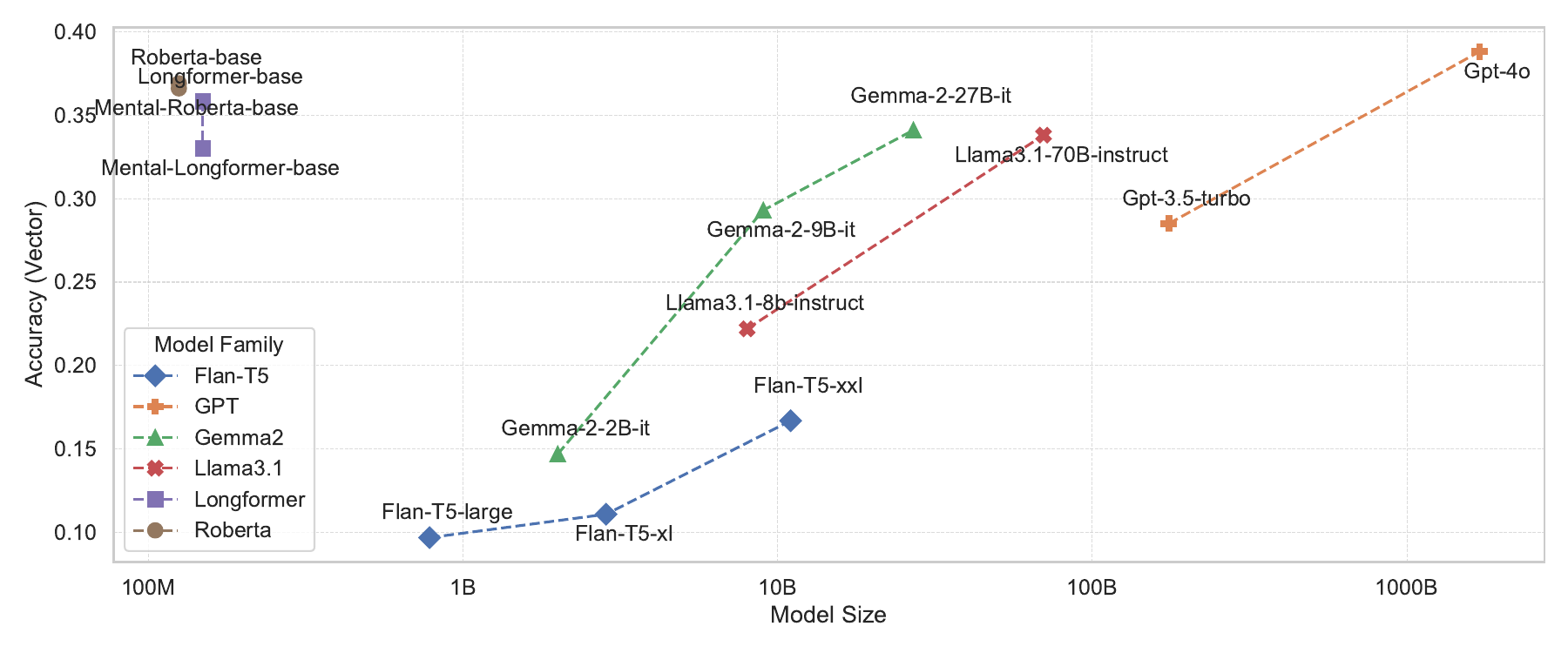}
    \caption{A comparison of model size, family, and emotion detection vector accuracy. The results show that model performance is significantly influenced by the model family and generally improves with increasing model size across four causal language model families.}
    \label{fig:accuracy_vs_models}
\end{figure*}

\noindent\textbf{Model Family and Size Matter.} Although the tested language models do not achieve either $Acc_L$ or $Acc_V$ higher than 0.4 under zero-shot settings, the emotion recognition performance varies significantly between models, as shown in Figure~\ref{fig:accuracy_vs_models}. Three factors appear to have the most significant impact on performance: \textbf{Model Architecture}, \textbf{Model Family}, and \textbf{Model Size}.

The four masked language models are all ranked among the top seven models in both accuracy metrics, despite having far fewer parameters (RoBERTa with 125M and Longformer with 149M). This may be due to their specialization in mask-filling tasks: they are specifically designed and trained to fill masks and do not need to interpret complex prompts as Seq2seq or causal language models do. In Seq2seq and causal language models, the model family plays a crucial role. For example, with similar model sizes, Flan-T5-xxl, Llama3.1-8B, and Gemma-2-9B exhibit vastly different $Acc_L$, $Acc_V$, and $F1_V$ scores. Flan-T5-xxl and Gemma2-9B have differences of 0.156, 0.174, and 0.135 in $Acc_L$, $Acc_V$, and $F1_V$, respectively. GPT-3.5-turbo, despite having 175B parameters, performs worse than smaller models such as Llama-3.1-70B and Gemma-2-27B.

Another key factor influencing performance is model size. Within all model families—Flan-T5, Llama-3.1, Gemma-2, and GPTs—the ability to predict emotions improves consistently as the number of parameters increases, without exception. Interestingly, the best-performing causal language model, GPT-4o, with 1.75T parameters, performs similarly to the three masked language models. This highlights the significant gap in mask-filling ability between masked language models and causal language models in emotion recognition tasks. The Wilcoxon Signed-Rank tests were conducted, and the results are shown in Table~\ref{tab:statistical_modeling_2}.

\noindent\textbf{Chain of Thought Doesn't Help Models Predict Emotions.} Studies show that CoT prompting enables models to reason step by step to arrive at the final answer and has been proven to improve performance on various NLP tasks~\cite{wei2022chain}. Given that most emotional experience texts in our dataset are long (259 words per post on average), leading models to reason step by step might be a potential way to achieve higher performance. To test this, we adapted CoT prompting by instructing models to "Think step by step"~\cite{kojima2023largelanguagemodelszeroshot}. The prompts we used are shown in the Appendix (Table~\ref{tab:combined-prompt}).

However, as Table~\ref{tab:results} shows, for all seven tested models across three model families and various sizes, performance consistently worsened, with average decreases of 0.039, 0.037, and 0.026 for $Acc_L$, $Acc_V$, and $F1_V$, respectively. The only models unaffected were GPT-4o and Llama-3.1-8B. Notably, for the three Gemma-2 models, regardless of size, the models' ability to respond in the correct format as instructed deteriorated significantly, dropping from an average of 99.9\% to 76\%. The Wilcoxon Signed-Rank tests were conducted and the results show that all models, except for these two, perform statistically significantly worse under CoT settings $ (p < 0.001) $. For smaller models, such as Gemma-2-9B, CoT prompting sometimes caused them to deviate from the original instruction and fail to respond with emotion words. For larger models, while CoT prompting did not reduce the rate of valid responses, overall performance declined. This aligns with prior findings~\cite{shaikh-etal-2023-second}, suggesting that CoT may underperform in socially situated tasks. It also echoes recent work~\cite{chochlakis2024largerlanguagemodelsdont}, which found that CoT fails to improve outcomes in complex, context-sensitive, and subjective tasks, particularly for larger models, which tend to rely heavily on their built-in prior knowledge, potentially leading them to overlook or disregard the specific context provided in the prompt.

\noindent\textbf{Error Analysis.}
To better understand the details of the prediction errors, we conducted an error analysis on the best-performing causal language models: Llama-3.1-70B-Instruct, Gemma-2-27B-It, and GPT-4o. We observed a significant difference between the distribution of emotion words used in human self-disclosures and those predicted by the models. Humans tend to use emotion words such as happy, scared, sad, tired, and embarrassed, whereas the models frequently overused emotion words such as anxious, grateful, overwhelmed, ashamed, frustrated, and relieved, as shown in Figure~\ref{fig:label_pred_count} in the Appendix.

Among the most common mispredictions, errors frequently occur when the model predicts a similar but distinct emotion, or one with a different intensity. For example, models often predict grateful, a deeper and more enduring emotional expression, when the true label is thankful. They also tend to overuse words like frustrated and anxious to represent a wide range of emotional experiences, whereas humans express these emotions more diversely and subtly in these cases using words like disheartened, demoralized, irritated, annoyed, restless, panicked, agitated, and stressed. Additionally, models often reduce emotional intensity by predicting angry instead of furious, or afraid instead of terrified. Table~\ref{tab:normalized-errors} in the Appendix presents the normalized mispredictions commonly made by these three models.

\noindent\textbf{Impact of Dataset Segmentation on Model Performance}
When constructing EXPRESS, we segmented long posts into segments of 512 tokens to ensure a fair and consistent evaluation across LLMs with different context length. However, this segmentation may limit the ability of LLMs with extended context windows, like GPT-4o to fully utilize their contextual reasoning capabilities, potentially underestimating their performance. To assess this trade-off, we conducted an additional analysis comparing model performance on segmented versus full posts. We randomly sampled 1,000 posts from the EXPRESS and evaluated GPT-4o, our best-performing model with a 128k-token context window, on both settings.

As Table~\ref{tab:segmentation_comparison} shows, despite a substantial increase in post length (approximately 1,000 additional words), performance remained effectively unchanged across all metrics. This suggests that a 512-token context window provides sufficient context for models to make comparable decisions on this task. Post segmentation not only ensures a fair and consistent evaluation across LLMs with different context lengths, but also does not disadvantage models with larger context capacities in this setting.

\begin{table}[ht]
\centering
\fontsize{7}{7}\selectfont
\begin{tabular}{lcccc}
\toprule
\textbf{Setting} & \textbf{Avg Post Length (words)} & \textbf{AccL} & \textbf{AccV} & \textbf{F1V} \\
\midrule
Segmented Post & 353  & 0.348 & 0.404 & 0.717 \\
Full Post      & 1353 & 0.346 & 0.406 & 0.715 \\
\bottomrule
\end{tabular}
\caption{Comparison of GPT-4o’s performance on segmented vs. full posts.}
\label{tab:segmentation_comparison}
\end{table}

\section{Are Models Good Learners of Emotions?}

Our results in the previous section indicate that accurately predicting self-disclosed emotions based on emotional experiences remains a challenge for language models, both at the lexical level and the basic emotion vector level. Even the best-performing models, including the four masked language models, Llama-3.1-70B-Instruct, Gemma-2-27B-Instruct, and GPT-4o, incorrectly predict at least one basic emotion dimension in the emotion vector for nearly 65\% of instances.

However, most language models are neither designed nor trained specifically for emotional intelligence tasks, such as emotion recognition. As a result, they may lack the implicit capacity to predict emotions in a human-like manner. Therefore, it is crucial to examine whether LLMs can learn and improve their emotion detection capabilities. Demonstrating this potential would highlight their utility in assisting mental health-related tasks, particularly when specifically designed and trained for such applications. To evaluate this, we established two few-shot experimental settings.

The two few-shot settings use different strategies to select examples. The first setting is designed to examine whether LLMs can learn from random examples of emotions, serving as a baseline. The second few-shot setting is designed to assess whether LLMs learn better when provided with examples of similar emotional experiences. To achieve this, we use the BERT-base-uncased model~\cite{devlin2019bert} to compute sentence embeddings and apply cosine similarity to identify the four nearest examples to the test query as few-shot examples.

\begin{figure}[t]
    \centering
    \includegraphics[width=0.36\textwidth]{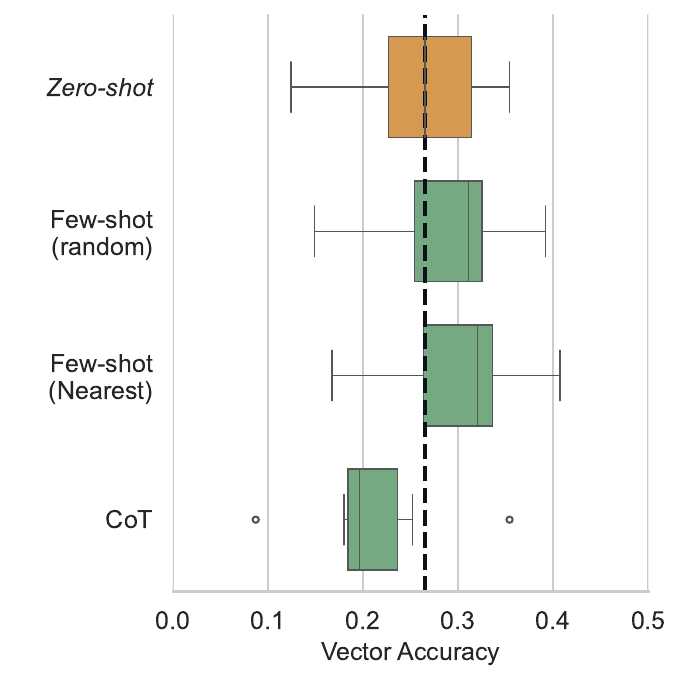}
    \caption{A comparison of zero-shot results and the other three settings. Models learn from examples and perform better compared to the zero-shot setting.}
    \label{fig:settings_vs_zero_shot_single}
\end{figure}

\noindent\textbf{Result} As shown in Table~\ref{tab:results} and Figure~\ref{fig:settings_vs_zero_shot_single}, all tested LLMs demonstrate improved emotion recognition ability when exposed to random examples. The average improvements in $Acc_L$, $Acc_V$, and $F1_V$ are 0.032, 0.042, and 0.006, respectively. Similar to, but even better than, the first few-shot setting, the few-shot setting with the nearest distanced examples further boosts the models’ performance on emotion recognition. This setting outperforms the first setting for all models, with average improvements in $Acc_L$, $Acc_V$, and $F1_V$ of 0.045, 0.053, and 0.011, respectively, compared to the zero-shot setting. This indicates that models can learn better when exposed to examples with similar emotional experiences. The improved results of the two few-shot settings demonstrate the ability of models to learn from provided examples, indicating their potential to become more emotionally aware in future training and adapt to mental health applications.

\section{Human Evaluation of Predicted and Actual Emotions}

To contextualize our empirical evaluation of LMs’ emotion recognition capabilities across the previous experiments, we conducted a qualitative analysis of a subset of the predicted (by LLMs) and actual emotions. Our first research objective was to \textit{assess which emotion (predicted/actual) would be picked by a domain emotion expert as being more plausible, or consistent with academic theory and empirical research about emotion experience}. 

Our second research question was: \textit{“How might a domain expert in emotion research determine that a certain emotion fits the natural contexts in the posts  (e.g., select an LLM-generated vs. ground truth emotion as being accurate)?”} Essentially, we aim to compute a form of \textbf{contextual accuracy} to assess whether the predicted or ground truth emotion aligns with the local context of the text. Hence, we perform a qualitative analysis on a small subset of our dataset to compute this contextual accuracy, utilizing human evaluation by three emotion experts, whose backgrounds are detailed in the Human Evaluation section of the Appendix.

We framed this task as a Turing test \cite{turing2009computing}, aiming to see if an LLM can mimic human emotional intelligence by attempting to deceive the emotion experts into selecting the LLM's output over the self-disclosed emotion. To maximize the challenge, we utilize the best-performing model’s outputs (GPT-4o with few-shot-nearest setting) for this task.

\subsection{Setup}

Our sample includes 213 EXPRESS posts, the associated self-disclosed emotions, and the corresponding predictions from the best-performing model. We selected the posts where the emotions predicted by the LLM differ from the self-disclosed emotions at the vector level. We randomized the order of presenting predicted or actual emotions first, and asked the coders to read each post and select one of the following options without knowing which options were predicted or actual: (1) self-disclosed emotion (SD), (2) LLM-generated emotion (LLM), (3) both emotions equally fit (BOTH), and (4) neither emotion fits (NEITHER). For clarity, it is important to note that selecting one emotion does not necessarily imply that the other option is unsuitable; rather, it indicates that the chosen emotion is preferred by the coders.

In the next phase, one coder went through all samples and provided rationales for their selections. Open coding~\cite{khandkar2009open} was then conducted on the rationales to understand what the common themes were to justify the selections made, as presented in Table~\ref{tab:human_eval_codes} in Appendix. The codes were then presented to the other coders, who were also free to add additional codes; however, no additional codes were introduced.

\subsection{Results}

Two coders selected LLM over SD more frequently, while the third coder selected SD more often. We used a majority vote to aggregate the results from the three coders. Overall, SD was selected 89 times (40.0\%), whereas LLM was selected 97 times (43.7\%). The coders also considered BOTH emotions plausible 5 times (4.7\%), and there were 22 instances (11.6\%) where all three chose different options. However, given the subtle distinctions between emotions, agreement among the coders was relatively low, with only 36.2\% of the samples receiving unanimous selection and a Fleiss’ Kappa score of 0.21 (fair agreement).

We aggregated the codes from the three coders and reported the frequency of codes in Table~\ref{tab:human_eval_codes} in the Appendix. The most common reason for coders selecting the SD emotions was that the selection better matched other terms providing context for the affective experience (Code 3, N=101). The second most common reason was that the selection was judged as better fitting the degree of specificity in the affective experience being described (Code 2, N=39). For the LLM emotions, the most common reasons were similar (Code 2, N=69; Code 3, N=58, and Code 5, N=57). However, when choosing SD, Code 3 was significantly more frequent (36.4\%), indicating that coders believed SD emotions were more contextually appropriate in many cases. In contrast, when choosing LLM, Code 2 and Code 5 were significantly more frequent (23\% and 19\%), suggesting that coders preferred LLM-generated emotions due to their higher specificity in describing the affective experience and better alignment with emotion definitions and theories. Examples of code selections and the corresponding rationales are presented in Table~\ref{tab:emotion_qual_codes}.

Our results unexpectedly show that experts slightly preferred the LLM's predicted emotions over the self-disclosed emotions. The low agreement between coders further reveals that both SD and LLM-generated emotions are plausible, differing primarily in nuanced ways. This finding highlights the inherent complexity and subtlety of our task, where human understanding and expression of emotions are highly subjective, shaped by an individual’s unique experiences, personality, emotion granularity, and perspective. This pattern of findings, across the three coders, suggests that while our best-performing LLM is able to generate emotions consistent with theoretical definitions and convincing as being appropriate, there are also instances where LLM may miss important contextual cues in the excerpts that may seem intuitive to human coders. However, it is important to note that these findings are specific to the best-performing model under optimal settings, which may represent the upper bound of LLM performance. Further investigation is needed to explore whether, in such cases where a human coder can articulate how an important clue in an excerpt is "missed" by the LLM, the LLM is making "errors" (such as by not accounting for certain information) or registering the contextual information but making predictions based on different information within the excerpt, potentially relying on "internal world modeling" specific to the LLM.

\section{Conclusion}

Higher emotional understanding and prediction abilities in LLMs are crucial for empathetic interactions~\cite{mayer1996emotional}, as users often prefer models that align with their beliefs~\cite{kirk2024benefits}. Misjudging or failing to respond empathetically in dialogue can lead to user discomfort~\cite{ball2000emotion}. However, existing benchmarks are limited by unreliable labels, a narrow range of emotion categories, and short emotional contexts.

To address these gaps, we constructed EXPRESS, an emotion recognition dataset created by masking self-disclosed emotions in Reddit posts. Our systematic evaluation revealed that LLMs still face challenges in aligning with human emotional expressions. Performance varied across model architectures and families, with consistent improvements as model size increased. Notably, while model performance varied among model families, masked language models performed comparably to some larger causal language models, such as GPT-4o. Given their significantly smaller model sizes, they offer a cost-effective alternative with similar performance.

We also tested whether CoT prompting improves LLM performance. Our findings, consistent with prior work~\cite{shaikh-etal-2023-second,chochlakis2024largerlanguagemodelsdont}, show that CoT degrades performance in subjective tasks, possibly due to models relying too heavily on prior knowledge instead of contextual cues. Few-shot prompting, however, showed promise, indicating that with targeted design and exposure, LLMs can improve their emotion recognition performance even without explicit emotion-task training.

For the qualitative analysis, future research could explore how the affective terms predicted by LLMs differ from self-disclosed emotions and how expert observations might be used to fine-tune models for improved emotion recognition. The analysis also shows that LLM-generated emotions were preferred by experts half of the time, suggesting that GPT-4o, under the best settings, has the ability to generate reasonable emotions that fit the context. However, the error analysis and qualitative analysis reveal that while LLM emotions are sometimes more specific than SD emotions and consistent with emotion theories, they can also be overly general, frequently predicting common emotions such as `anxious' or `frustrated.' Moreover, LLMs are sometimes less effective at capturing contextual cues than SD emotions. These findings, along with the low alignment with SD emotions, highlight the importance of self-disclosed emotions in fine-grained emotion recognition tasks, which not only serve as a benchmark for evaluation but also provide valuable training material to improve model alignment.

Our benchmark and evaluation framework offer a systematic way to assess LLMs' capacity to understand and predict fine-grained, self-disclosed emotions. While our setup is intentionally controlled, it provides foundational insights into how models handle nuanced emotional expressions, a prerequisite for deployment in sensitive, real-world applications such as mental health support tools or social media moderation. As these applications demand accurate emotion recognition, models that underperform in our benchmark may require further careful examination before being reliably applied in such contexts. 

\section{Limitations}

\noindent \textbf{Diversity in emotion expression.} While our research provides valuable insights into emotion recognition, it is primarily focused on neurotypical ways of expressing emotions. This limitation highlights the need for further research to explore and understand how emotional expressions may differ in neurodivergent populations, as differences in emotional expression are well-documented~\cite{trevisan2017adults}. Expanding the scope of future studies to include a more diverse range of emotional expressions will help create more inclusive models and improve the accuracy and applicability of emotion recognition systems across different populations~\cite{mazefsky2012need}.

\noindent \textbf{Bias in User Demographics.} Reddit's user base is not fully representative of the general population. Studies have shown that Reddit users are predominantly male, young adults, with a strong representation from North America and Europe~\cite{barthel2016reddit,singer2014evolution}. This demographic bias may influence the types of emotions expressed and the language used on the platform.
Thus, fine-tuning models on EXPRESS may not generalize well to other populations, leading to potential biases in the emotion recognition model when applied to more diverse or global datasets. Future work could broaden the demographic coverage by incorporating data from other platforms, such as Quora, or region-specific platforms like Zhihu (Chinese) in other languages. Future work can also oversample specific subreddits to target different demographic groups, such as r/askwomen for women and r/over60 for older adults.

\noindent \textbf{Limitations to Human Evaluation Approach. } There are few, if any, real-world scenarios where a person would be tasked with predicting the emotion term that an emotion-experiencer would express via a written vignette. Rather, real-world scenarios involve some discussion and clarifying questions for an individual to learn about what a person might be feeling or have experienced in the past (e.g., emotional disclosure to a friend, or a therapist). A task similar to the one in this study is a subscale of an emotional intelligence measure in which participants determine what emotion a person is feeling based on a scenario described. However, this task uses standardized vignettes developed by researchers, rather than disclosed in colloquial terms by everyday people. As suggested by the instances where both the self-disclosed and LLM terms were considered by the expert as plausible colloquial descriptions of affect, there may be cases where a term cannot be predicted by a human (at least without more context provided). Furthermore, future work could include further refinement through iterative codebook development and discussions to improve agreement among coders. 

\section{Ethics Statement}

Emotion detection ability in LLMs could bring significant benefits to mental health applications. It would allow for the automation of mental health services, directing individuals to appropriate and personalized resources. This approach could enhance the accessibility of mental health support, particularly for vulnerable populations, such as ethnic minorities, where seeking help is often more stigmatized compared to majority groups \cite{stade2024large,habicht2024closing}. However, there are potential risks as well. With increased emotional intelligence, large language models might become more persuasive, increasing the potential to manipulate vulnerable populations. Additionally, LLMs tend to be sycophantic \cite{sharma2023towards}, and their emotional intelligence may lead them to better align with a human’s opinions and sentiments. This personalization carries risks, as it can reinforce the user's existing beliefs \cite{kirk2024benefits}. Consequently, LLMs may avoid suggesting mental health resources that fall outside the user’s comfort zone, instead adhering to assumptions based on the user's prompts and selectively presenting information that reflects the user’s biases and beliefs.

The annotations used to obtain the basic emotion vectors for Plutchik's eight basic emotions were crowd-sourced, with workers receiving \$0.10 per assignment, ensuring that compensation complied with the minimum wage requirements in the authors' location. Each HIT allowed sufficient time of 3 minutes for completion, aligned with the number of questions included. To ensure quality, only workers with a HIT approval rate of 95\% or higher, at least 5,000 approved HITs, and who passed a task-specific qualification test were allowed to perform the annotations. We also recruited one domain expert on Upwork for qualitative analysis, compensating the expert at \$30 per hour, with the entire evaluation process taking 11 hours. To maintain data anonymity, we discarded post IDs and account names before feeding the posts into the LLMs. Our dataset is self-annotated, strictly extracting emotions explicitly expressed by the post author.

\bibliography{aaai25,anthology}

\begin{thebibliography}{72}
\providecommand{\natexlab}[1]{#1}

\bibitem[{Achiam et~al.(2023)Achiam, Adler, Agarwal, Aleman et~al.}]{achiam2023gpt}
Achiam, J.; Adler, S.; Agarwal, S.; Aleman, F.~L.; et~al. 2023.
\newblock Gpt-4 technical report.
\newblock \emph{arXiv preprint arXiv:2303.08774}.

\bibitem[{Adhikary et~al.(2024)Adhikary, Srivastava, Kumar, Singh, Manuja et~al.}]{adhikary2024exploring}
Adhikary, P.~K.; Srivastava, A.; Kumar, S.; Singh, S.~M.; Manuja, P.; et~al. 2024.
\newblock Exploring the Efficacy of Large Language Models in Summarizing Mental Health Counseling Sessions: Benchmark Study.
\newblock \emph{JMIR Mental Health}, 11: e57306.

\bibitem[{Alm and Sproat(2005)}]{Alm2005PerceptionsOE}
Alm, C.~O.; and Sproat, R. 2005.
\newblock Perceptions of Emotions in Expressive Storytelling.
\newblock In \emph{INTERSPEECH}, volume 2005, 533--536.

\bibitem[{Ball and Breese(2001)}]{ball2000emotion}
Ball, G.; and Breese, J. 2001.
\newblock \emph{Emotion and personality in a conversational agent}, 189–219.
\newblock Cambridge, MA, USA: MIT Press.
\newblock ISBN 0262032783.

\bibitem[{Barthel et~al.(2016)Barthel, Stocking, Holcomb, and Mitchell}]{barthel2016reddit}
Barthel, M.; Stocking, G.; Holcomb, J.; and Mitchell, A. 2016.
\newblock Reddit news users more likely to be male, young and digital in their news preferences.
\newblock \emph{Pew Research Center}, 25.

\bibitem[{Beltagy, Peters, and Cohan(2020)}]{beltagy2020longformerlongdocumenttransformer}
Beltagy, I.; Peters, M.~E.; and Cohan, A. 2020.
\newblock Longformer: The Long-Document Transformer.
\newblock arXiv:2004.05150.

\bibitem[{Boe(2021)}]{redditPRAW}
Boe, B. 2021.
\newblock {PRAW}: The {P}ython {R}eddit {API} {W}rapper.
\newblock \url{praw.readthedocs.io/en/latest/}.

\bibitem[{Brackett, Rivers, and Salovey(2011)}]{brackett2011emotional}
Brackett, M.~A.; Rivers, S.~E.; and Salovey, P. 2011.
\newblock Emotional intelligence: Implications for personal, social, academic, and workplace success.
\newblock \emph{Social and personality psychology compass}, 5(1): 88--103.

\bibitem[{Brown et~al.(2020)Brown, Mann, Ryder et~al.}]{brown2020language}
Brown, T.; Mann, B.; Ryder, P., Nick~Dhariwal; et~al. 2020.
\newblock Language models are few-shot learners.
\newblock \emph{Advances in neural information processing systems}, 33: 1877--1901.

\bibitem[{Buechel and Hahn(2017)}]{buechel2022emobank}
Buechel, S.; and Hahn, U. 2017.
\newblock {E}mo{B}ank: Studying the Impact of Annotation Perspective and Representation Format on Dimensional Emotion Analysis.
\newblock In Lapata, M.; Blunsom, P.; and Koller, A., eds., \emph{Proceedings of the 15th Conference of the {E}uropean Chapter of the Association for Computational Linguistics: Volume 2, Short Papers}, 578--585.

\bibitem[{Chatterjee et~al.(2019)Chatterjee, Narahari, Joshi, and Agrawal}]{chatterjee2019semeval}
Chatterjee, A.; Narahari, K.~N.; Joshi, M.; and Agrawal, P. 2019.
\newblock SemEval-2019 task 3: EmoContext contextual emotion detection in text.
\newblock In \emph{Proceedings of the 13th international workshop on semantic evaluation}, 39--48.

\bibitem[{Chochlakis et~al.(2025)Chochlakis, Pandiyan, Lerman, and Narayanan}]{chochlakis2024largerlanguagemodelsdont}
Chochlakis, G.; Pandiyan, N.~M.; Lerman, K.; and Narayanan, S. 2025.
\newblock Larger Language Models Don’t Care How You Think: Why Chain-of-Thought Prompting Fails in Subjective Tasks.
\newblock In \emph{ICASSP 2025 - 2025 IEEE International Conference on Acoustics, Speech and Signal Processing (ICASSP)}, 1--5.

\bibitem[{Chung et~al.(2024)Chung, Hou, Longpre et~al.}]{chung2022flant5}
Chung, H.~W.; Hou, L.; Longpre, J., Shayne~Wei; et~al. 2024.
\newblock Scaling instruction-finetuned language models.
\newblock \emph{J. Mach. Learn. Res.}, 25(1).

\bibitem[{Chung and Harris(2018)}]{mingi2018jealousy}
Chung, M.; and Harris, C.~R. 2018.
\newblock Jealousy as a Specific Emotion: The Dynamic Functional Model.
\newblock \emph{Emotion Review}, 10(4): 272--287.

\bibitem[{Cowen et~al.(2019)Cowen, Sauter, Tracy, and Keltner}]{cowen2019}
Cowen, A.; Sauter, D.; Tracy, J.~L.; and Keltner, D. 2019.
\newblock Mapping the Passions: Toward a High-Dimensional Taxonomy of Emotional Experience and Expression.
\newblock \emph{Psychological Science in the Public Interest}, 20(1): 69--90.
\newblock PMID: 31313637.

\bibitem[{Davis(2024)}]{berkeleywellbeingListEmotions}
Davis, T. 2024.
\newblock {L}ist of {E}motions: 271 {E}motion {W}ords (+ {P}{D}{F}) --- berkeleywellbeing.com.
\newblock \url{https://www.berkeleywellbeing.com/list-of-emotions.html}.
\newblock [Accessed 06-06-2024].

\bibitem[{Davitz(2013)}]{davitz2013language}
Davitz, J. 2013.
\newblock \emph{The Language of Emotion}.
\newblock Academic Press.
\newblock ISBN 9781483261713.

\bibitem[{Demszky et~al.(2020)Demszky, Movshovitz-Attias, Ko, Cowen, Nemade, and Ravi}]{demszky2020goemotions}
Demszky, D.; Movshovitz-Attias, D.; Ko, J.; Cowen, A.; Nemade, G.; and Ravi, S. 2020.
\newblock {G}o{E}motions: A Dataset of Fine-Grained Emotions.
\newblock In \emph{Proceedings of the 58th Annual Meeting of the Association for Computational Linguistics}, 4040--4054.

\bibitem[{Devlin et~al.(2019)Devlin, Chang, Lee, and Toutanova}]{devlin2019bert}
Devlin, J.; Chang, M.-W.; Lee, K.; and Toutanova, K. 2019.
\newblock {BERT}: Pre-training of Deep Bidirectional Transformers for Language Understanding.
\newblock In \emph{Proceedings of the 2019 Conference of the North {A}merican Chapter of the Association for Computational Linguistics: Human Language Technologies, Volume 1 (Long and Short Papers)}, 4171--4186.

\bibitem[{Ekman, Sorenson, and Friesen(1969)}]{ekman1969pan}
Ekman, P.; Sorenson, E.~R.; and Friesen, W.~V. 1969.
\newblock Pan-cultural elements in facial displays of emotion.
\newblock \emph{Science}, 164(3875): 86--88.

\bibitem[{Ekman et~al.(1999)}]{ekman1999basic}
Ekman, P.; et~al. 1999.
\newblock Basic emotions.
\newblock \emph{Handbook of cognition and emotion}, 98(45-60): 16.

\bibitem[{Feng et~al.(2022)Feng, Lubis, Geishauser, Lin, Heck, van Niekerk, and Gasic}]{feng2021emowoz}
Feng, S.; Lubis, N.; Geishauser, C.; Lin, H.-c.; Heck, M.; van Niekerk, C.; and Gasic, M. 2022.
\newblock {E}mo{WOZ}: A Large-Scale Corpus and Labelling Scheme for Emotion Recognition in Task-Oriented Dialogue Systems.
\newblock In \emph{Proceedings of the Thirteenth Language Resources and Evaluation Conference}, 4096--4113.

\bibitem[{Fontaine et~al.(2007)Fontaine, Scherer, Roesch, and Ellsworth}]{fontaine2007nottwo}
Fontaine, J.~R.; Scherer, K.~R.; Roesch, E.~B.; and Ellsworth, P.~C. 2007.
\newblock The World of Emotions is not Two-Dimensional.
\newblock \emph{Psychological Science}, 18(12): 1050--1057.
\newblock PMID: 18031411.

\bibitem[{Frattaroli(2006)}]{frattaroli2006experimental}
Frattaroli, J. 2006.
\newblock Experimental Disclosure and Its Moderators: A Meta-Analysis.
\newblock \emph{Psychological Bulletin}, 132(6): 823.

\bibitem[{Ghazi, Inkpen, and Szpakowicz(2015)}]{ghazi2015detecting}
Ghazi, D.; Inkpen, D.; and Szpakowicz, S. 2015.
\newblock Detecting emotion stimuli in emotion-bearing sentences.
\newblock In \emph{Computational Linguistics and Intelligent Text Processing: 16th International Conference, CICLing 2015, Cairo, Egypt, April 14-20, 2015, Proceedings, Part II 16}, 152--165. Springer.

\bibitem[{Grattafiori et~al.(2024)Grattafiori, Dubey, Jauhri et~al.}]{grattafiori2024llama3herdmodels}
Grattafiori, A.; Dubey, A.; Jauhri, A.; et~al. 2024.
\newblock The Llama 3 Herd of Models.
\newblock arXiv:2407.21783.

\bibitem[{Grootendorst(2022)}]{grootendorst2022bertopic}
Grootendorst, M. 2022.
\newblock BERTopic: Neural topic modeling with a class-based TF-IDF procedure.
\newblock \emph{arXiv preprint arXiv:2203.05794}.

\bibitem[{Gross and Feldman~Barrett(2011)}]{gross2011emotion}
Gross, J.~J.; and Feldman~Barrett, L. 2011.
\newblock Emotion generation and emotion regulation: One or two depends on your point of view.
\newblock \emph{Emotion review}, 3(1): 8--16.

\bibitem[{Habicht et~al.(2024)Habicht, Viswanathan, Carrington, Hauser, Harper, and Rollwage}]{habicht2024closing}
Habicht, J.; Viswanathan, S.; Carrington, B.; Hauser, T.~U.; Harper, R.; and Rollwage, M. 2024.
\newblock Closing the accessibility gap to mental health treatment with a personalized self-referral Chatbot.
\newblock \emph{Nature medicine}, 30(2): 595--602.

\bibitem[{Harmon-Jones, Price, and Gable(2012)}]{harmon2012influence}
Harmon-Jones, E.; Price, T.~F.; and Gable, P.~A. 2012.
\newblock The Influence of Affective States on Cognitive Broadening/Narrowing: Considering the Importance of Motivational Intensity.
\newblock \emph{Social and Personality Psychology Compass}, 6(4): 314--327.

\bibitem[{Hua et~al.(2025)Hua, Na, Li, Liu, Fang, Clifton, and Torous}]{hua2024large}
Hua, Y.; Na, H.; Li, Z.; Liu, F.; Fang, X.; Clifton, D.; and Torous, J. 2025.
\newblock A scoping review of large language models for generative tasks in mental health care.
\newblock \emph{npj Digital Medicine}, 8(1): 230.

\bibitem[{Ji et~al.(2022)Ji, Zhang, Ansari, Fu, Tiwari, and Cambria}]{ji2022mentalbert}
Ji, S.; Zhang, T.; Ansari, L.; Fu, J.; Tiwari, P.; and Cambria, E. 2022.
\newblock MentalBERT: Publicly Available Pretrained Language Models for Mental Healthcare.
\newblock In \emph{Proceedings of the Thirteenth Language Resources and Evaluation Conference}, 7184--7190.

\bibitem[{Ji et~al.(2023)Ji, Zhang, Yang, Ananiadou, Cambria, and Tiedemann}]{ji2023mentallongformer}
Ji, S.; Zhang, T.; Yang, K.; Ananiadou, S.; Cambria, E.; and Tiedemann, J. 2023.
\newblock Domain-specific Continued Pretraining of Language Models for Capturing Long Context in Mental Health.
\newblock \emph{arXiv preprint arXiv:2304.10447}.

\bibitem[{Jurafsky and Martin(2024)}]{dan2024}
Jurafsky, D.; and Martin, J.~H. 2024.
\newblock \emph{Speech and Language Processing: An Introduction to Natural Language Processing, Computational Linguistics, and Speech Recognition with Language Models}.
\newblock 3rd edition.
\newblock Online manuscript released August 20, 2024.

\bibitem[{Khandkar(2009)}]{khandkar2009open}
Khandkar, S.~H. 2009.
\newblock Open coding.
\newblock \emph{University of Calgary}, 23(2009): 2009.

\bibitem[{Kirk et~al.(2024)Kirk, Vidgen, R{\"o}ttger, and Hale}]{kirk2024benefits}
Kirk, H.~R.; Vidgen, B.; R{\"o}ttger, P.; and Hale, S.~A. 2024.
\newblock The benefits, risks and bounds of personalizing the alignment of large language models to individuals.
\newblock \emph{Nature Machine Intelligence}, 1--10.

\bibitem[{Kojima et~al.(2022)Kojima, Gu, Reid, Matsuo, and Iwasawa}]{kojima2023largelanguagemodelszeroshot}
Kojima, T.; Gu, S.~S.; Reid, M.; Matsuo, Y.; and Iwasawa, Y. 2022.
\newblock Large language models are zero-shot reasoners.
\newblock In \emph{Proceedings of the 36th International Conference on Neural Information Processing Systems}, NIPS '22. Red Hook, NY, USA.
\newblock ISBN 9781713871088.

\bibitem[{Li et~al.(2017)Li, Su, Shen, Li, Cao, and Niu}]{li2017dailydialog}
Li, Y.; Su, H.; Shen, X.; Li, W.; Cao, Z.; and Niu, S. 2017.
\newblock {D}aily{D}ialog: A Manually Labelled Multi-turn Dialogue Dataset.
\newblock In \emph{Proceedings of the Eighth International Joint Conference on Natural Language Processing (Volume 1: Long Papers)}, 986--995.

\bibitem[{Liu, Osama, and De~Andrade(2019)}]{liu2019dens}
Liu, C.; Osama, M.; and De~Andrade, A. 2019.
\newblock {DENS}: A Dataset for Multi-class Emotion Analysis.
\newblock In \emph{Proceedings of the 2019 Conference on Empirical Methods in Natural Language Processing and the 9th International Joint Conference on Natural Language Processing (EMNLP-IJCNLP)}, 6293--6298.

\bibitem[{Liu et~al.(2021{\natexlab{a}})Liu, Shen, Zhang, Dolan, Carin, and Chen}]{liu2021makes}
Liu, J.; Shen, D.; Zhang, Y.; Dolan, B.; Carin, L.; and Chen, W. 2021{\natexlab{a}}.
\newblock What Makes Good In-Context Examples for GPT-$3 $?
\newblock \emph{arXiv preprint arXiv:2101.06804}.

\bibitem[{Liu et~al.(2021{\natexlab{b}})Liu, Zheng, Demasi, Sabour, Li, Yu, Jiang, and Huang}]{liu-etal-2021-towards}
Liu, S.; Zheng, C.; Demasi, O.; Sabour, S.; Li, Y.; Yu, Z.; Jiang, Y.; and Huang, M. 2021{\natexlab{b}}.
\newblock Towards Emotional Support Dialog Systems.
\newblock In Zong, C.; Xia, F.; Li, W.; and Navigli, R., eds., \emph{Proceedings of the 59th Annual Meeting of the Association for Computational Linguistics and the 11th International Joint Conference on Natural Language Processing (Volume 1: Long Papers)}, 3469--3483. Online: Association for Computational Linguistics.

\bibitem[{Liu et~al.(2019)Liu, Ott, Goyal, Du, Joshi, Chen, Levy, Lewis, Zettlemoyer, and Stoyanov}]{liu2019robertarobustlyoptimizedbert}
Liu, Y.; Ott, M.; Goyal, N.; Du, J.; Joshi, M.; Chen, D.; Levy, O.; Lewis, M.; Zettlemoyer, L.; and Stoyanov, V. 2019.
\newblock RoBERTa: A Robustly Optimized BERT Pretraining Approach.
\newblock arXiv:1907.11692.

\bibitem[{Mauss and Robinson(2009)}]{mauss2009measures}
Mauss, I.~B.; and Robinson, M.~D. 2009.
\newblock Measures of Emotion: A Review.
\newblock \emph{Cognition and Emotion}, 23(2): 209--237.

\bibitem[{Mayer and Geher(1996)}]{mayer1996emotional}
Mayer, J.~D.; and Geher, G. 1996.
\newblock Emotional intelligence and the identification of emotion.
\newblock \emph{Intelligence}, 22(2): 89--113.

\bibitem[{Mazefsky, Pelphrey, and Dahl(2012)}]{mazefsky2012need}
Mazefsky, C.~A.; Pelphrey, K.~A.; and Dahl, R.~E. 2012.
\newblock The need for a broader approach to emotion regulation research in autism.
\newblock \emph{Child development perspectives}, 6(1): 92--97.

\bibitem[{Min et~al.(2022)Min, Lyu, Holtzman, Artetxe, Lewis, Hajishirzi, and Zettlemoyer}]{min2022rethinking}
Min, S.; Lyu, X.; Holtzman, A.; Artetxe, M.; Lewis, M.; Hajishirzi, H.; and Zettlemoyer, L. 2022.
\newblock Rethinking the Role of Demonstrations: What Makes In-Context Learning Work?
\newblock In \emph{Proceedings of the 2022 Conference on Empirical Methods in Natural Language Processing}, 11048--11064.

\bibitem[{Mohammad(2018)}]{mohammad2018obtaining}
Mohammad, S. 2018.
\newblock Obtaining reliable human ratings of valence, arousal, and dominance for 20,000 English words.
\newblock In \emph{Proceedings of the 56th annual meeting of the association for computational linguistics (volume 1: Long papers)}, 174--184.

\bibitem[{Mohammad and Bravo-Marquez(2017)}]{mohammad2017emotion}
Mohammad, S.; and Bravo-Marquez, F. 2017.
\newblock Emotion Intensities in Tweets.
\newblock In Ide, N.; Herbelot, A.; and M{\`a}rquez, L., eds., \emph{Proceedings of the 6th Joint Conference on Lexical and Computational Semantics (*{SEM} 2017)}, 65--77.

\bibitem[{Mohammad and Turney(2013)}]{mohammad2013crowdsourcing}
Mohammad, S.~M.; and Turney, P.~D. 2013.
\newblock Crowdsourcing a word--emotion association lexicon.
\newblock \emph{Computational intelligence}, 29(3): 436--465.

\bibitem[{{\"O}hman et~al.(2018){\"O}hman, Kajava, Tiedemann, and Honkela}]{ohman2018creating}
{\"O}hman, E.; Kajava, K.; Tiedemann, J.; and Honkela, T. 2018.
\newblock Creating a dataset for multilingual fine-grained emotion-detection using gamification-based annotation.
\newblock In \emph{Proceedings of the 9th workshop on computational approaches to subjectivity, sentiment and social media analysis}, 24--30.

\bibitem[{OpenAI et~al.(2024)OpenAI, Achiam, Adler et~al.}]{openai2024gpt4technicalreport}
OpenAI; Achiam, J.; Adler, S.; et~al. 2024.
\newblock GPT-4 Technical Report.
\newblock arXiv:2303.08774.

\bibitem[{Pennebaker and Beall(1986)}]{pennebaker1986confronting}
Pennebaker, J.~W.; and Beall, S.~K. 1986.
\newblock Confronting a Traumatic Event: Toward an Understanding of Inhibition and Disease.
\newblock \emph{Journal of Abnormal Psychology}, 95(3): 274.

\bibitem[{Plutchik(1980)}]{plutchik1980general}
Plutchik, R. 1980.
\newblock A general psychoevolutionary theory of emotion.
\newblock In \emph{Theories of emotion}, 3--33. Elsevier.

\bibitem[{Russell(1980)}]{russell1980circumplex}
Russell, J.~A. 1980.
\newblock A circumplex model of affect.
\newblock \emph{Journal of personality and social psychology}, 39(6): 1161.

\bibitem[{Sabour et~al.(2024)Sabour, Liu, Zhang, Liu, Zhou, Sunaryo, Lee, Mihalcea, and Huang}]{sabour2024emobench}
Sabour, S.; Liu, S.; Zhang, Z.; Liu, J.; Zhou, J.; Sunaryo, A.; Lee, T.; Mihalcea, R.; and Huang, M. 2024.
\newblock {E}mo{B}ench: Evaluating the Emotional Intelligence of Large Language Models.
\newblock In \emph{Proceedings of the 62nd Annual Meeting of the Association for Computational Linguistics (Volume 1: Long Papers)}, 5986--6004.

\bibitem[{Salovey and Mayer(1990)}]{salovey1990emotional}
Salovey, P.; and Mayer, J.~D. 1990.
\newblock Emotional intelligence.
\newblock \emph{Imagination, cognition and personality}, 9(3): 185--211.

\bibitem[{Scherer and Wallbott(1994)}]{scherer1994evidence}
Scherer, K.~R.; and Wallbott, H.~G. 1994.
\newblock Evidence for universality and cultural variation of differential emotion response patterning.
\newblock \emph{Journal of personality and social psychology}, 66(2): 310.

\bibitem[{Schiller et~al.(2024)Schiller, Alessandra, Alia-Klein, Dolcos et~al.}]{schiller2024human}
Schiller, D.; Alessandra, N.; Alia-Klein, N.; Dolcos, F.; et~al. 2024.
\newblock The human affectome.
\newblock \emph{Neuroscience \& Biobehavioral Reviews}, 158: 105450.

\bibitem[{Shaikh et~al.(2023)Shaikh, Zhang, Held, Bernstein, and Yang}]{shaikh-etal-2023-second}
Shaikh, O.; Zhang, H.; Held, W.; Bernstein, M.; and Yang, D. 2023.
\newblock On Second Thought, Let`s Not Think Step by Step! Bias and Toxicity in Zero-Shot Reasoning.
\newblock In Rogers, A.; Boyd-Graber, J.; and Okazaki, N., eds., \emph{Proceedings of the 61st Annual Meeting of the Association for Computational Linguistics (Volume 1: Long Papers)}, 4454--4470. Toronto, Canada: Association for Computational Linguistics.

\bibitem[{Sharma et~al.(2024)Sharma, Tong, Bowman et~al.}]{sharma2023towards}
Sharma, M.; Tong, M.; Bowman, S.~R.; et~al. 2024.
\newblock Towards Understanding Sycophancy in Language Models.
\newblock In \emph{The Twelfth International Conference on Learning Representations}.

\bibitem[{Siemer, Mauss, and Gross(2007)}]{siemer2007same}
Siemer, M.; Mauss, I.; and Gross, J.~J. 2007.
\newblock Same Situation--Different Emotions: How Appraisals Shape Our Emotions.
\newblock \emph{Emotion (Washington, D.C.)}, 7(3): 592--600.

\bibitem[{Singer et~al.(2014)Singer, Fl{\"o}ck, Meinhart, Zeitfogel, and Strohmaier}]{singer2014evolution}
Singer, P.; Fl{\"o}ck, F.; Meinhart, C.; Zeitfogel, E.; and Strohmaier, M. 2014.
\newblock Evolution of reddit: from the front page of the internet to a self-referential community?
\newblock In \emph{Proceedings of the 23rd international conference on world wide web}, 517--522.

\bibitem[{Singh, Caragea, and Li(2024)}]{singh2023language}
Singh, S.; Caragea, C.; and Li, J.~J. 2024.
\newblock Language Models (Mostly) Do Not Consider Emotion Triggers When Predicting Emotion.
\newblock In \emph{Proceedings of the 2024 Conference of the North American Chapter of the Association for Computational Linguistics: Human Language Technologies (Volume 2: Short Papers)}, 603--614.

\bibitem[{Stade et~al.(2024)Stade, Stirman, Ungar, Boland, Schwartz, Yaden, Sedoc, DeRubeis, Willer, and Eichstaedt}]{stade2024large}
Stade, E.~C.; Stirman, S.~W.; Ungar, L.~H.; Boland, C.~L.; Schwartz, H.~A.; Yaden, D.~B.; Sedoc, J.; DeRubeis, R.~J.; Willer, R.; and Eichstaedt, J.~C. 2024.
\newblock Large language models could change the future of behavioral healthcare: a proposal for responsible development and evaluation.
\newblock \emph{NPJ Mental Health Research}, 3(1): 12.

\bibitem[{Strapparava and Mihalcea(2007)}]{strapparava2007semeval}
Strapparava, C.; and Mihalcea, R. 2007.
\newblock Semeval-2007 task 14: Affective text.
\newblock In \emph{Proceedings of the fourth international workshop on semantic evaluations (SemEval-2007)}, 70--74.

\bibitem[{Tangney et~al.(1996)Tangney, Miller, Flicker, and Barlow}]{Tangney1996shame}
Tangney, J.~P.; Miller, R.~S.; Flicker, L.; and Barlow, D.~H. 1996.
\newblock Are shame, guilt, and embarrassment distinct emotions?
\newblock \emph{Journal of Personality and Social Psychology}, 70: 1256--1269.

\bibitem[{Team et~al.(2024)Team, Riviere, Pathak et~al.}]{gemmateam2024gemma2improvingopen}
Team, G.; Riviere, M.; Pathak, S.; et~al. 2024.
\newblock Gemma 2: Improving Open Language Models at a Practical Size.
\newblock arXiv:2408.00118.

\bibitem[{Trevisan et~al.(2017)Trevisan, Roberts, Lin, and Birmingham}]{trevisan2017adults}
Trevisan, D.~A.; Roberts, N.; Lin, C.; and Birmingham, E. 2017.
\newblock How do adults and teens with self-declared Autism Spectrum Disorder experience eye contact? A qualitative analysis of first-hand accounts.
\newblock \emph{PloS one}, 12(11): e0188446.

\bibitem[{Turing(2009)}]{turing2009computing}
Turing, A.~M. 2009.
\newblock \emph{Computing machinery and intelligence}.
\newblock Springer.

\bibitem[{Van~Pelt and Sorokin(2012)}]{van2012designing}
Van~Pelt, C.; and Sorokin, A. 2012.
\newblock Designing a scalable crowdsourcing platform.
\newblock In \emph{Proceedings of the 2012 ACM SIGMOD International Conference on Management of Data}, 765--766.

\bibitem[{Wang et~al.(2023)Wang, Li, Yin, Wu, and Liu}]{wang2023emotional}
Wang, X.; Li, X.; Yin, Z.; Wu, Y.; and Liu, J. 2023.
\newblock Emotional intelligence of large language models.
\newblock \emph{Journal of Pacific Rim Psychology}, 17: 18344909231213958.

\bibitem[{Wei et~al.(2022)Wei, Wang, Schuurmans, Zhou et~al.}]{wei2022chain}
Wei, J.; Wang, X.; Schuurmans, D.; Zhou, D.; et~al. 2022.
\newblock Chain-of-thought prompting elicits reasoning in large language models.
\newblock \emph{Advances in neural information processing systems}, 35: 24824--24837.

\end{thebibliography}
\appendix
\section{Appendix}

\renewcommand{\thefigure}{A.\arabic{figure}}
\renewcommand{\thetable}{A.\arabic{table}}
\setcounter{figure}{0}
\setcounter{table}{0}

\subsection{Theories of Emotion}

\noindent There are several popular theories of emotion. One class of theories views emotions as being discrete, otherwise known as basic emotions. Plutchik's wheel of emotion \cite{plutchik1980general} consists of 8 discrete emotions: \textit{joy, sadness, anger, fear, trust, disgust, anticipation,} and \textit{surprise}. Another similar theory is Ekman's basic emotions \cite{ekman1969pan}, consisting of 6 basic emotions: \textit{happiness, anger, fear, sadness, disgust,} and \textit{surprise}. On the other hand, some theories view emotions in a multi-dimensional space, such as Russel's Circumplex model of affect~\cite{russell1980circumplex}. The VAD model places emotions among dimensions of valence, arousal, and dominance. Valence, arousal, and dominance are dimensions used to describe emotions: valence indicates the positivity or negativity of emotion, arousal reflects the intensity of emotional activation, and dominance measures the degree of control or influence an emotion exerts over an individual~\cite{russell1980circumplex}.

Many works use these theories as a basis to create lexicons for emotion. Popular lexicons include the NRC Word-Emotion Association Lexicon, also known as EmoLex \cite{mohammad2013crowdsourcing}, and the NRC Valence, Arousal, and Dominance lexicon \cite{mohammad2018obtaining}.

\subsection{Emotion Masking Algorithm}
We included the pattern `feel' because prior work indicated that humans use the word feeling' interchangeably with emotion, even though feelings and emotions are not the same~\cite{berkeleywellbeingListEmotions}. Feelings encompass both emotional experiences (e.g., feeling sad) and physical sensations (e.g., feeling hungry). This distinction justifies our pattern-matching format (\textit{I + feel/am + emotion}) or (\textit{no-pronoun + feeling + emotion}). Up to three words can be added between `\textit{I}' and `\textit{feel/am}', and between `\textit{feel/am}' and the emotion, allowing the patterns to capture cases where extra words such as adverbs are present. For example, the word `angry' in the phrase `I have felt extremely angry' will be masked.

In addition to these basic patterns, several guidelines were introduced to ensure the accuracy of masking self-disclosed emotion words:

\begin{enumerate}
    \item We avoided masking the word if there is a pronoun, noun, or verb between `feel' and the emotion word, to exclude phrases like `I feel he was sad.'
    \item We avoided masking the word if there is an interrogative word between `feel' or `am' and the emotion word, to exclude phrases like `I feel how happy he is.'
    \item For the `I am' pattern, we ensured that the emotion word is an adjective.
    \item Finally, we performed a manual review to filter out posts that did not satisfy the conditions but were not detected by the protocol.
\end{enumerate}

\subsection{Post Segmentation}

We segment the posts into chunks of 512 tokens using RoBERTa tokenizer due to input context length constraints of RoBERTa and Longformer. Table~\ref{tab:algo} outlines the algorithm used to perform post segmentation.

\subsection{Evaluation Metrics}

The F1-score for a vector $\mathbf{v}_i$ is computed as:
\[
F1(\mathbf{v}_i) = \frac{2 \cdot \text{Precision}(\mathbf{v}_i) \cdot \text{Recall}(\mathbf{v}_i)}{\text{Precision}(\mathbf{v}_i) + \text{Recall}(\mathbf{v}_i)}
\]
where:
\[
\text{Precision}(\mathbf{v}_i) = \frac{\text{TP}_i}{\text{TP}_i + \text{FP}_i}, \quad
\text{Recall}(\mathbf{v}_i) = \frac{\text{TP}_i}{\text{TP}_i + \text{FN}_i}
\]
Here, $\text{TP}_i$, $\text{FP}_i$, and $\text{FN}_i$ represent the true positives, false positives, and false negatives for the prediction of the ten dimensions in vector $\mathbf{v}_i$, respectively.

The final F1-score across all vectors is obtained by averaging the vector-level F1-scores:
\[
\text{Final F1} = \frac{1}{n} \sum_{i=1}^{n} F1(\mathbf{v}_i)
\]
where $n$ is the total number of vectors.

\begin{table*}[!htb]
\centering
\fontsize{8}{8}\selectfont
\begin{tabular}{|p{0.9\textwidth}|}
\hline
\textbf{Post Segmentation Algorithm} \\ \hline

\textbf{Step 1: Initialize grouping of [MASK] tokens} \\
\textbf{Step 1.1:} Traverse through the tokens of the post one by one. \\
\textbf{Step 1.2:} Whenever a [MASK] token is encountered, start a new group if it’s the first one, or add it to the current group if it's the first [MASK] token found. \\ \hline

\textbf{Step 2: Group nearby [MASK] tokens} \\
\textbf{Step 2.1:} Check the distance between the current [MASK] token and the last [MASK] token in the current group. \\
\textbf{Step 2.2:} If the distance is less than or equal to 235 tokens, add this [MASK] token to the current group. \\
\textbf{Step 2.3:} Else, finalize the current group and start a new group with this [MASK] token. \\ \hline

\textbf{Step 3: Compute centroid for each [MASK] group} \\
For each group of [MASK] tokens, calculate the mean position by averaging the positions of all [MASK] tokens in that group. \\ \hline

\textbf{Step 4: Create segments around centroids} \\
For each centroid, create a segment by selecting tokens around this central position. Take \textit{n/2} tokens to the left and \textit{n/2} tokens to the right of the centroid to create a segment of \textit{n} tokens in total (510 tokens). \\ \hline

\textbf{Step 5: Clip the segments} \\
Ensure that each segment makes sense contextually by clipping the segment to sentence boundaries. Clip the segment slightly to align with the nearest sentence-ending characters. \\ \hline

\textbf{Step 6: Return the segments} \\
Once all [MASK] token groups have been segmented and clipped, return the list of these segments as the output. \\ \hline

\end{tabular}
\caption{Details the algorithm for clipping the dataset into 510-token segments}
\label{tab:algo}
\end{table*}

\begin{table*}[!htb]
\centering
\fontsize{8}{8}\selectfont
\begin{tabular}{@{}l|r|r|r|r@{}}
    \hline
     & mean & median & min & max\\
    \hline
    Words per Post & $ 259 ~(\pm 132.26)$ & $281$  & 20 & 492\\
    Emotion words per Post & $ 1.56 ~(\pm 0.87)$ & $1$ & 1 & 5\\
    \hline
\end{tabular}
\caption{Statistics of the Reddit posts in our dataset}
\label{table:dataset_stats}
\end{table*}

\begin{table*}[!htb]
\centering
  \scriptsize
  \begin{tabular}{@{}|p{0.8\textwidth}|p{0.1\textwidth}}
    \toprule
    \textbf{Text} & \textbf{Label} \\
    \midrule
    My family's absolutely wonderful traditional Japanese New Year feast. Soy sauce *everywhere*, but this year I felt \texttt{<mask>} to just be with everyone, instead of sad I can't eat. They even made me separate rice balls with plain seaweed - sometimes family gatherings are stressful but this was great!  & grateful \\
    \midrule
    Today I decided to post this bare face photo to Snapchat instead of using a filter and actually preferred how I looked without it  I dont have perfect skin, but I officially feel more \texttt{<mask>} without any makeup than I do with. & confident \\
    \midrule
    Show this page on Pinterest the other day, and was instantly eager to recreate it on my journal, i hope i did justice. Also, on the right side is my playlist that i made last night with few of my favourite mood lifters and i feel <mask> about it. Love to hear your feedbacks! :). & happy \\
    \midrule
    Decided to draw while high last night... then proceeded to write 8 pages trying to prove a 4th spatial dimension. I feel \texttt{<mask>} and like this picture represents a dark part of my emotions. & enlightened \\
    \midrule
    My mom has been super into canning lately and was excited for summer fruits and veggies, but she broke her wrist last week and had to have surgery. I feel so \texttt{<mask>} for her. Any ideas about how to support her canning passion while she recovers? & sad \\
    \midrule
    The rest of the gallery erupted in cheers and applause as the judge handed down the death sentence and I too felt a wave of \texttt{<mask>} that this monster would face justice. I recognized the fairness of the court, but hounded by an agonizing regret I also wondered where I went wrong and longed bitterly for a do-over for the little boy my son had once been. & relief \\
    \midrule
    I feel so <mask> when there's no one beside me, as long as someone is there, a friend or a boyfriend nor a group of friends. I feel good and <mask>. The minute am alone it starts to feel lonely, like I have no one in the world.  & lonely, cheerful\\
    \midrule
    I just finished a two-year solo project and I feel <mask> and depressed at the same time. I'm playing with my app now ... it's alive! A ton of research, scrounging, all night coding around family life. Learned a ton, aged a lot but it was something I had to do after all those years of working on someone else's project. I'm so <mask> but I feel this weird sense of depression at the same time. A sort of loss. Since I'm a solo keyboard warrior, I guess you guys are the ones with whom I sharing this weird feeling(s). & elated, happy\\
    \midrule
    Lately I feel immense <mask> for my life and I wish I knew who to thank. Lately I feel overwhelmingly <mask> for the good things that have come into my life. I even feel <mask> for some of the bad things that have happened to me in the last couple years, because without those I wouldn't be a person ready to accept the good things that are happening lately. I don't believe in God but I quite often throughout my life feel that there is some energy looking out for me. Someone who has my best interest at heart and has the wisdom and ability to nudge me in the directions I need to go. Not always the directions I want to go, but always the ones I need. I wish I knew who that was, so I could thank them.
    & gratitude, grateful, grateful\\
    \bottomrule
  \end{tabular}
  \caption{Examples from EXPRESS}
  \label{tab:dataset_examples_appendix}
\end{table*}

\begin{table}[h]
\centering
\fontsize{8}{8}\selectfont
\begin{tabular}{ll ll ll}
\toprule
\textbf{Emotion} & \textbf{Count} & \textbf{Emotion} & \textbf{Count} & \textbf{Emotion} & \textbf{Count} \\
\midrule
happy        & 2999 & guilty       & 2778 & scared      & 2637 \\
sad          & 2212 & tired        & 1853 & afraid      & 1624 \\
anxious      & 1405 & comfortable  & 1115 & embarrassed & 1113 \\
angry        & 1107 & worried      & 1100 & depressed   & 1078 \\
lonely       & 1055 & grateful     & 991  & confused    & 954  \\
excited      & 924  & nervous      & 894  & confident   & 857  \\
upset        & 826  & overwhelmed  & 798  & uncomfortable & 776 \\
proud        & 668  & numb         & 645  & ashamed     & 619  \\
hopeless     & 596  & surprised    & 522  & uneasy      & 501  \\
guilt        & 471  & helpless     & 462  & frustrated  & 458  \\
interested   & 456  & hurt         & 454  & humiliated  & 424  \\
thankful     & 414  & disgusted    & 394  & disappointed& 386  \\
miserable    & 370  & mad          & 363  & terrified   & 344  \\
hopeful      & 329  & love         & 326  & shocked     & 319  \\
stuck        & 313  & isolated     & 289  & jealous     & 283  \\
restless     & 277  & relief       & 275  & loved       & 254  \\
calm         & 248  & fear         & 248  & shame       & 243  \\
optimistic   & 235  & weak         & 234  & unsure      & 232  \\
anger        & 225  & relieved     & 217  & joy         & 213  \\
stressed     & 209  & bored        & 201  & anxiety     & 192  \\
\bottomrule
\end{tabular}
\caption{Top 60 most frequent emotion lexicons in the EXPRESS dataset.}
\label{tab:top60_emotions}
\end{table}

\subsection{Topic Modeling}
\begin{figure*}[!htb]
  \centering
  \includegraphics[scale=0.26]{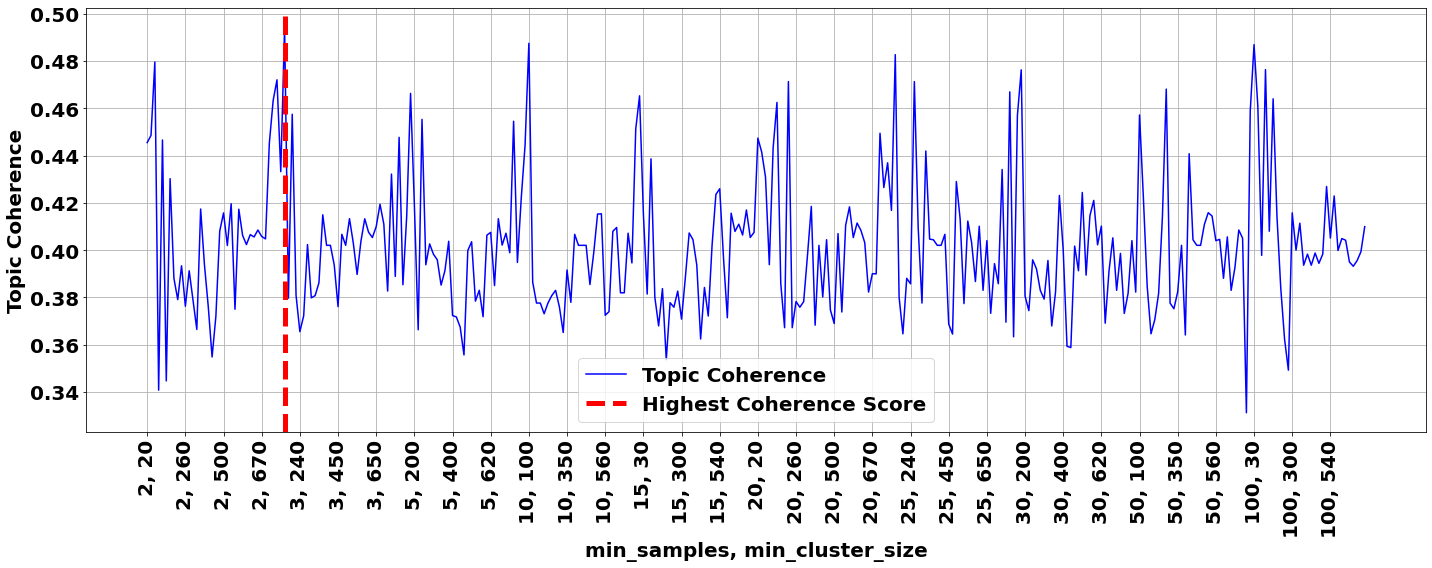}
  \caption{The topic coherence graph illustrates the coherence scores for all the hyperparameter combinations tested. The highest coherence score achieved is 0.494. The x-axis represents the different hyperparameter combination pairs.}
  \label{fig:topic_coh}
\end{figure*}

We use BERTopic to perform topic modeling on our dataset. BERTopic leverages document embeddings, reducing their dimensionality before clustering them \cite{grootendorst2022bertopic}. To optimize our model's performance, we conducted hyperparameter tuning on the HDBSCAN algorithm.\footnote{We experiment with two hyperparameters: \\ min\_sample = [2, 3, 5, 10, 15, 20, 25, 30, 50, 100] \\ min\_cluster\_size = [20, 25, 30, 50, 100, 150, 200, 220, 240, 250, 260, 280, 300, 330, 350, 380, 400, 420, 450, 460, 500, 520, 540, 550, 560, 600, 620, 640, 650, 660, 670, 700]} The highest coherence score ($C_v$) achieved was 0.494, with a min\_sample value of 3 and a min\_cluster\_size of 100, as illustrated in Figure~\ref{fig:topic_coh}. This configuration resulted in the identification of 49 distinct topics, which are detailed in Table~\ref{tab:topic_labels}, including the top four representative words and qualitative labels assigned to each topic.


We conducted similar topic modeling on other datasets if they did not report topic diversity and presented the results in Table~\ref{tab:lit_review}.


\begin{table*}[!htb]
\scalebox{1}{
\fontsize{8}{8}\selectfont
\begin{tabular}{l|l}
\multicolumn{1}{l|}{\textbf{Topic ID and Name}}                             & \textbf{Qualitative label}      \\ \hline
 -1\_serenejsan\_disgustawe\_\_       & no topic               \\
 0\_like\_friend\_dont\_know              & relationships          \\
1\_back\_could\_eye\_one                 & sex and pregnancy \textit{(sex)}                   \\
 2\_job\_work\_year\_like                 & new career or career transition \textit{(career)}                   \\
3\_like\_day\_trip\_time                 & drugs and mental illness \textit{(drugs)}                  \\
4\_game\_player\_play\_character         & gaming                 \\
5\_woman\_men\_trans\_gay                & sexuality and identity \\
6\_weight\_lb\_mile\_race                & health and weight loss \\
7\_god\_church\_jesus\_christian         & christianity           \\
8\_makeup\_comfortable\_confident\_dress & self-confidence and body image \textit{(self-confidence)}       \\
9\_hair\_skin\_picking\_acne             & personal appearance    \\
10\_anxiety\_anxious\_panic\_attack      & anxiety                \\
11\_husband\_wife\_ha\_told              & marriage               \\
12\_grief\_died\_mom\_life               & death                  \\
13\_dog\_cat\_vet\_puppy                 & pets                   \\
14\_meditation\_experience\_like\_life   & meditation             \\
15\_film\_movie\_character\_season       & movies                 \\
16\_team\_optimistic\_player\_league     & sports                 \\
17\_anger\_angry\_like\_emotion          & anger                  \\
18\_worthy\_art\_posting\_drawing        & art                    \\
19\_pain\_symptom\_doctor\_day           & medical symptoms and issues \textit{(medical symptoms)}         \\
20\_song\_album\_music\_uzi              & music                  \\
21\_sad\_cheer\_today\_depressed         & sadness and loneliness \textit{(loneliness)}                \\
22\_patient\_nurse\_surgery\_hospital    & surgery and medical procedures \textit{(medical procedures)}             \\
23\_happy\_today\_joy\_happiness         & happiness              \\
24\_excited\_excitement\_tender\_enjoy   & excitement             \\
25\_people\_like\_dont\_think            & social isolation       \\
26\_school\_class\_teacher\_student      & school                 \\
27\_drinking\_drink\_sober\_alcohol      & alcoholism             \\
28\_embarrassed\_humiliated\_like\_today & embarrassment           \\
29\_grateful\_gratitude\_today\_thankful & gratitude              \\
30\_fear\_scared\_tara\_douma            & fear                   \\
31\_guilty\_guilt\_like\_dont            & guilt                  \\
32\_porn\_sex\_sexual\_lust              & sexual and porn addiction \textit{(sexual addiction)}                    \\
33\_vegan\_meat\_food\_eat               & veganism               \\
34\_wedding\_sister\_family\_friend      & weddings               \\
35\_confident\_today\_lb\_confidence     & confidence due to weight loss \textit{(confidence)}             \\
36\_sleep\_bed\_wake\_night              & sleep                  \\
37\_dream\_like\_nightmare\_woke         & dreams and nightmares  \\
38\_pride\_shame\_proud\_sense           & pride and shame        \\
39\_ocd\_thought\_like\_intrusive        & OCD                    \\
40\_book\_ron\_read\_character           & books                  \\
41\_ako\_house\_lang\_hector             & housing              \\
42\_insulted\_offended\_insult\_coach    & insult                 \\
43\_birthday\_gift\_today\_friend        & birthdays              \\
44\_affection\_affectionate\_love\_hug   & love                   \\
45\_lust\_passion\_lustful\_passionate   & lust                   \\
46\_manager\_office\_customer\_work      & workplace dynamics              \\
47\_happiness\_happy\_joy\_life          & happiness and depression              \\
48\_agony\_despair\_fazgoo\_anguish      & pain                  
\end{tabular}
}
\caption{The top 4 words for each topic generated through topic modeling and their respective qualitative labels.}
\label{tab:topic_labels}
\end{table*}

\begin{table}[ht]
\centering
\fontsize{8}{8}\selectfont
\begin{tabular}{lll}
\toprule
\textbf{Label} & \textbf{Prediction} & \textbf{Frequency} \\
\midrule
thankful     & grateful     & 0.60 \\
disheartened & frustrated   & 0.33 \\
demoralized  & frustrated   & 0.32 \\
eager        & excited      & 0.31 \\
furious      & angry        & 0.29 \\
pleased      & proud        & 0.29 \\
optimistic   & hopeful      & 0.27 \\
terrified    & afraid       & 0.27 \\
restless     & anxious      & 0.25 \\
scared       & afraid       & 0.24 \\
irritated    & frustrated   & 0.24 \\
panicked     & anxious      & 0.24 \\
agitated     & anxious      & 0.23 \\
powerless    & helpless     & 0.23 \\
thrilled     & excited      & 0.23 \\
stress       & anxiety      & 0.23 \\
annoyed      & frustrated   & 0.23 \\
disoriented  & overwhelmed  & 0.23 \\
impatient    & frustrated   & 0.22 \\
passionate   & excited      & 0.21 \\
\bottomrule
\end{tabular}
\caption{Top normalized mispredictions (vector level) by top three causal language models.}
\label{tab:normalized-errors}
\end{table}

\subsection{Human Evaluation}

\noindent \textbf{Coder Background.}  
All three of our coders are domain experts with Ph.D. degrees in related fields:

\begin{itemize}
    \item \textbf{Coder One:} A domain expert with a Ph.D. in Psychological Science, specializing in Affective Science, with over 10 years of research experience using a combination of quantitative surveys, qualitative interviews, and experimental designs to understand emotional experience and regulation, as well as lay beliefs and academic theories about emotion.

    \item \textbf{Coder Two:} A domain expert with a Ph.D. in Psychology and Social Behavior, with multiple peer-reviewed publications related to emotion.

    \item \textbf{Coder Three:} A domain expert with a Ph.D. in Counseling Psychology. They have extensive experience working with diverse adult populations in various clinical settings, including VA Medical Centers. Their therapeutic approach emphasizes two main roles: providing empathetic support for clients' immediate concerns and fostering deeper self-awareness of internal and interpersonal patterns that may contribute to distress or dissatisfaction. They also have experience teaching as an adjunct instructor at major universities.

\end{itemize}

\begin{table*}[!htb]
\fontsize{7}{7}\selectfont
\begin{tabular}{@{}|p{0.04\textwidth}|p{0.34\textwidth}|p{0.54\textwidth}|}
\cline{1-3}
\textbf{Category} & \textbf{Qualitative Code Description} & \textbf{Example} \\ \cline{1-3}

Applies to SD and LLM   

& Code 1 ($N_{SD}=24, N_{LLM}=19$): Valence (positive/pleasant or negative/unpleasant) matched the context of the situation and/or the affective reaction described (N=18) &  My sister just suffered an ectopic pregnancy. I feel guilty for being thankful that it happened before her state banned abortion. Yesterday my pregnant sister was rushed to the ER with excruciating pain. She was 7 weeks pregnant. Once they realized she had an ectopic pregnancy, the doctors gave her the medicine to stop it. Luckily, my sister is fine, nothing ruptured. I am <mask> but also livid - what if this happened four weeks later? What if the doctors didn't give her the proper care, so my sister had to endure the pain and potentially put her life at risk? What if it causes her to never have children again? All because of the Supreme Courts f*cked up decision to overturn Roe v Wade. \newline \textit{\textcolor{darkgreen}{label: sad}, \textcolor{red}{\underline{predicted: relieved}}}\\ 
\cline{2-3}

& Code 2 ($N_{SD}=39, N_{LLM}=69$): Terms have similar valence (positive/pleasant or negative/unpleasant) but the selection was judged as better fitting the degree of specificity in the affective experience being described (more general vs. more specific) & Am I a bad friend for feeling <mask> that my friend is replacing me? for the longest time, ive been really close to one of my friends. we always gossip together, go everywhere together, and do so many things together. unfortunately shes struggled with making friends for a really long time but recently shes been doing really well and making new friends and ive felt really proud of her. a friend of mine was hanging out with the both of us and the friend of mine had brought another friend. the friend of the friend has instantly clicked with my close friend and they were almost inseparable afterwards. i feel like i shouldnt  feel sad that shes getting closer to other people but she doesnt really talk to me when the other girl is there. \newline \textit{\textcolor{darkgreen}{label: upset}, \textcolor{red}{\underline{predicted: jealous}}}\\ 
\cline{2-3}

& Code 3 ($N_{SD}=101, N_{LLM}=58$): Selection better matches other term(s) providing context of the the affective experience &  Feel hatred for family sometimes. Like, when I say I'm <mask> or depressed. They tell me "then change it" like there's some f*cking switch you can flip and it all gets better. I'm working my body and brain into dust just to make rent. It ain't like the movies folks, the people closest to you cut the deepest \newline \textit{\textcolor{darkgreen}{\underline{label: unhappy}}, \textcolor{red}{predicted: frustrated}}\\ 
\cline{2-3}

& Code 4 ($N_{SD}=20, N_{LLM}=37$): Degree of intensity (greater or lesser) of the selection is better matched to or reflective of typical responses to the scenario described   & I feel so <mask>. I need advice please. I was in school I I've gotten a weird feeling like I was about to throw up, well I did not but there was this loud sound of like choking or something (?) I just heard people giggle in the back what do i do now? Please give me some advice i feel horrible right now \newline \textit{\textcolor{darkgreen}{label: humiliated}, \textcolor{red}{\underline{predicted: embarrassed}}}\\ 
\cline{2-3}

& Code 5 ($N_{SD}=25, N_{LLM}=57$): Selection better fits a definition of the emotion described in the scenario, as proposed by a peer-reviewed published theory/definition of the emotion  & I feel riddled with <mask> about whether hrt will be illegal within the near future, how can I deal with this? The far right has been launching an war on trans people and gender affirming care providers and I'm afraid that within a few years or less than a year, it will no longer be legal for me to access hrt. If the right doesn't ban hrt via legislature, they will at least try to intimidate doctors away from prescribing it via violent threats and intimidate pharmacies away from distributing it. And I don't want to go back to the way I was before, hrt has done so much for my happiness and confidence. But now it's going to be taken away from me again. And it causes me a ton of anxiety and sadness not knowing if I'll be forced to detransition someday soon. \newline \textit{\textcolor{darkgreen}{\underline{label: fear}}, \textcolor{red}{predicted: anxiety}}\\ 
\cline{2-3}

& Code 6 ($N_{SD}=16, N_{LLM}=16$): Selection reflects whether the event/scenario described has already occurred (post-goal emotion, e.g., happiness/joy or down) or is likely to occur (pre-goal emotion, e.g., excitement or nervous) & Feeling so <mask>. I (27F) posted in here a few weeks ago about the proposal that wasnt. Ive already been clear with my (30M) boyfriend on my timeline and hes on board. I just feel so taken for granted knowing that he has nothing planned for the foreseeable future. (Our weekends are all pretty much booked and spoken for from now until the end of the holiday season). Trying to shake this feeling of sadness and irritability whenever I see him. \newline \textit{\textcolor{darkgreen}{label: melancholy}, \textcolor{red}{\underline{predicted: disappointed}}}\\ \cline{2-3}

& Code 7 ($N_{SD}=9, N_{LLM}=11$): Alternative does not fit as well with evidence-based normative descriptions of affective experience in similar scenarios  & I feel <mask> to admit that i like the last of us part 2! i enjoyed the game and the story not at all bad! as youtubers and people made it seem. (according to me) and i enjoyed the gameplay and it felt good. the only thing weird was the >!dual protaganist!< but i had no problems with it! and the game looks so amazing im on a ps4 and its 60 fps too!i really enjoyed the game as much as i did the first one. \newline \textit{\textcolor{darkgreen}{label: scared}, \textcolor{red}{\underline{predicted: embarrassed}}} \\ 
\cline{2-3}

& Code 8 ($N_{SD}=18, N_{LLM}=13$): Alternative is not as consistent with colloquial descriptions of similar scenarios (according to the coder) & My parents refuse me medical attention for my depression because they believe God will heal me even though i attempted suicide. Help me make sense of it all. hello all.  Its not going good with me. My religious parents refuse me medical treatment for my depression and instead pray and force me to fast once a week. I tried to commit suicide last year November because i am messed up in life, in love with a black girl and they found out. My dad rambled racist stuff and i felt <mask> and lost.  she is the only thing i have so i overdosed. I have had mental issues for a year now and things compounded and i found my way in hospital.  My parents promised to get me helo. They didn't and change the topic. I am really suffering and God evem left me.  I am 15 and will become an atheist because of this. I already tore a Bible in anger.  Why would God allow this to happen and harden my parents heart and not even care.  My girlfriends family are waay better to be around and they all non religious. They not bigotic and treat me well. all efforts to help me are shut down by my parents. they tell me God is testing me and i must be strong and not weak.  help.  I think i will leave religion if rhis continues. \newline \textit{\textcolor{darkgreen}{\underline{label: hopeless}}, \textcolor{red}{predicted: lost}} \\ 
\cline{2-3}

& Code 9 ($N_{SD}=25, N_{LLM}=20$): Alternative does not logically follow as well as the alternative from the described situation, based on coder's lay theory of normative affective experience in similar situations   & Feeling <mask> after quiting weed. So I quit smoking weed at night 2 and a half weeks ago. I feel better and I'm not having any problems right now. But I feel so bored. I have stuff to do clean and cook, bake etc. I'm a stay at home mom and my husband works alot. I really want to play the Sims 3 on my laptop but I don't play it anymore because I used to be addicted to playing it. I'm figured I couldn't control my usage because of episodes before I was unmedicated and not diagnosed, but I'm stable now and think I can regulate how much I play but I've tried playing GTA on my ps4 and I still feel so <mask> idk if games would do anything for me. I don't feel like watching tv at all. I'm just here being bored all day. What do yall think what should I do? Has anyone been through this after stopping smoking weed? \newline \textit{\textcolor{darkgreen}{\underline{label: bored, bored}}, \textcolor{red}{predicted: anxious, disinterested}}\\ 
\cline{1-3}

Neither  & Code 10 ($N=4$): No information provided on the object/subject about which affect is being expressed (Unclear what the person is feeling affect about)  & Feeling <mask>. You guys are gonna have to deal with my attempts as I try to figure out face paint \textit{\textcolor{darkgreen}{label: melancholy}, \textcolor{red}{predicted: nervous}}\\ \cline{1-3}

Both  & Code 11 ($N=57$): Not enough information about the situation to determine whether on affect term was a better fit to the situation; rather, both terms could be applied such that the expression of affect would be equally or similarly believable & Just put my pc to sleep and felt very happy for some reason. Decided to take a picture. For absolutely no reason I felt very <mask> for having a computer like this. Its not the best out there but its mine. Does that happen to anyone else? \newline \textit{\textcolor{darkgreen}{\underline{label: thankful}}, \textcolor{red}{\underline{predicted: grateful}}}\\ 
\cline{1-3}

\end{tabular}

\caption{Descriptions of the qualitative codes developed by the emotion expert to categorize instances from the dataset. The underlined text indicates the option selected by the expert.}
\label{tab:human_eval_codes}
\end{table*}

\begin{table*}[!htb]
\fontsize{7}{7}\selectfont
\begin{tabular}{@{}|p{0.04\textwidth}|p{0.35\textwidth}|p{0.53\textwidth}|}
\hline
\textbf{Code} & \textbf{Selection and Reasoning} & \textbf{Example} \\ \hline

9 & Selection of hopeless (SD) instead of confused (LLM): based on the description of the situation indicating that the person appraises the situation as being bad, rather than unclear or confusing. 
& I feel <mask> and don't know what to do. To lay out a long story short.  I have been in a relationship for over six years…Within two months of my transferring, she cheated on me with another girl. She continued to talk to this girl despite me not wanting her to…This continued and she got more abusive again.  \\ \hline

9 & Selection of frustrated (LLM) instead of numb (SD): based on the coder’s lay theory that feeling the other emotions indicated in the excerpt would not constitute “numbness”, or lack of ability to feel.
& I'm <mask> because I got tired of seeing my mom and dads disappointing faces in everything I tried so hard at, just to fail. My parents have paid for multiple classes so I could actually be good at something but I never show any promise and It always ends up worse than when I started trying. I have one friend and I feel like hes my opposite…I'm starting to hate everything I once enjoyed because the thought of him being better than me at everything creeps into my mind. At this point I feel like there is no hope for me to be happy, just numb.  \\ \hline

9 & Selection of grateful (LLM) instead of happy (SD): based on the assumption that the person is sharing an account of less-than-ideal situation, rather than a situation to be happy for.
& Unfortunately I'm still low contact with them despite wanting to be no-contact (long story short, I got caught packing up my stuff and it turned into this huge f****** ordeal to the point where my online friend turned roommate who works a job involved with law enforcement had to drive 600 miles to rescue me and lowkey intimidated my family into letting me go), but GOD I will take the mental equivalent of weekly probation calls over the shit I was dealing with before. Hopefully I can fully ditch their asses someday. I'm just so <mask> to be able to exist, indulge myself in my hobbies and go to places without immediately being interrogated about what I'm doing or where I'm going or being. \\ \hline

2 & Selection of upset (SD) instead of jealous (LLM); based on lack of specificity of the alternative, that would be required to meet criteria for the definition of jealousy as a discrete emotion~\cite{mingi2018jealousy}.
& am i bad friend for feeling <mask> that my friend is replacing me?...i feel like i shouldnt  feel sad that shes getting closer to other people but she doesnt really talk to me when the other girl is there.  \\ \hline

4 & Selection of ashamed (LLM) instead of embarrassment (SD): given the use of the descriptor of feeling as being “so intense”, and based on findings that shame is relatively intense in terms of both negative valence, and physiological arousal, relative to embarrassment~\cite{Tangney1996shame}.
& …After the break up the feeling I felt was so intense and awful. Id just cry myself to sleep in silence because I knew no one understood how miserable I was and am. I never talk about it because Im <mask> I still care… \\ \hline

5 &  Selection of angry (SD) instead of frustrated (LLM): based important criteria met for the definition of anger as involving attribution of the other person as being at blame (rather than the self) for negative situation, in addition to goal blockage~\cite{siemer2007same}.
& …I messaged her again yesterday, she responded playing the blame game, wouldnt call me, said my family and I publicly posting that she fled was embarrassing her, were messed up for trying to hurt and embarrass her, etc. Basically blaming us for why she wouldnt reach out. I was <mask> and in tears. Classic behavior of someone struggling with addiction who wants to take no responsibility… This has been going on for years. I lost my adoptive father to alcohol and drugs. I sat with him as he was dying. I cannot watch my sister die. I cannot do this again. \\ \hline

6 & Selection of scared (SD) instead of guilty (LLM): based on the description of the goal (maintaining the status quo in family in the context of the father’s death) as being in future, rather than being in the past (attained, or not attained)~\cite{harmon2012influence}. 
& I feel like I want to die (not really), sort of, because I wanted so bad to be there for my Dad, who I adored, and I just couldn't do it…I loved  him so much, and I came back to the hospital after he passed, and held him forever, sobbing, telling him how much I missed him. He was so still, but I wasn't scared then for some reason.?   I think I was more <mask> that I was braver than my siblings, and wanted to stay, but didn't' want to upset the stupid family dynamic, when everyone was like "well I'm not staying, I can't take this, " and I felt compelled to leave as well, some stupid, weak show of solidarity.  I hate myself…  \\ \hline

\end{tabular}
\caption{Examples of selection and the reasoning for the selection of emotion terms, including corresponding codes and related theories for reference.}
\label{tab:emotion_qual_codes}
\end{table*}



\subsection{Prompt Design}

We manually tested multiple prompt variations, using accuracy score as a metric to assess performance improvements. Multiple prompt variations were tested, with accuracy score serving as the primary metric for evaluation. Our final prompt, which achieved the highest accuracy on our sample dataset, is detailed in Table~\ref{tab:combined-prompt}.

A critical consideration in our design was the definition of ``emotion''. We observed that providing an explicit definition of ``emotion'' often constrained the model's output, limiting it to a narrow set of predefined emotion terms. This conflicted with our objective of allowing the model to generate contextually appropriate emotion terms. To address this, we adopted a more flexible approach that avoids explicitly imposing a strict definition of emotions. This approach enables the model to capture nuanced emotional expressions without being restricted to predefined categories.

\begin{table*}[!htb]
\fontsize{7}{7}\selectfont
\centering
\begin{tabular}{@{}p{\textwidth}@{}}
\toprule
\textbf{Zero-shot Prompt:} \\
You are an assistant tasked with predicting emotion words masked as {<}mask{>} in a given self-disclosure text from social media. Predict the \{number of labels\} {<}mask{>} token\(s\) based on the context. \\
Provide your answer in the format\: {[}example\_format{]}. The length of the list must be \{number of labels\}. Only include words describing emotions, and provide no extra text or reasoning. \\
\\
\textbf{Text}: \{post\} \\
\textbf{Answer}: \\
\midrule

\textbf{Chain-of-Thought (CoT) Prompt:} \\
You are an assistant tasked with predicting emotion words masked as {<}mask{>} in a given self-disclosure text from social media. Predict the \{number of labels\} {<}mask{>} token\(s\) based on the context. \\
Think step by step to arrive at the final answer. In your response, first provide the reasoning, then provide your answer in the format: {[}example\_format{]}. The length of the list must be \{number of labels\}. Only include words describing emotions in the list. \\
\\
\textbf{Text}: \{post\} \\
\textbf{Answer}: \\
\midrule

\textbf{Few-shot Prompt:} \\
You are an assistant tasked with predicting emotion words masked as {<}mask{>} in a given self-disclosure text from social media. Predict the \{number of labels\} {<}mask{>} token\(s\) based on the context. \\
Provide your answer in the format\: {[}example\_format{]}. The length of the list must be \{number of labels\}. Only include words describing emotions, and provide no extra text or reasoning. \\
\\
\textbf{Examples:} \\
\textbf{Text}: $\{$ example\_text $\}$ \\
\textbf{Answer}: $\{$ example\_answer $\}$ \\
\textbf{Text}: $\{$ example\_text $\}$ \\
\textbf{Answer}: $\{$ example\_answer $\}$ \\
\textbf{Text}: $\{$ example\_text $\}$ \\
\textbf{Answer}: $\{$ example\_answer $\}$ \\
\textbf{Text}: $\{$ example\_text $\}$ \\
\textbf{Answer}: $\{$ example\_answer $\}$ \\
\\
Now, use this pattern for the given text. \\
\textbf{Text}: \{post\} \\
\textbf{Answer}: \\
\bottomrule
\end{tabular}
\caption{Prompt templates for Zero-shot, Chain-of-Thought (CoT), and Few-shot setups. \{number of labels\} and {[}example\_format{]} are adjustable based on the number of masks in the post.}
\label{tab:combined-prompt}
\end{table*}

\subsection{Amazon Mechanical Turk Annotation} 

\begin{figure*}[!htb]
  \centering
  \includegraphics[scale=0.21]{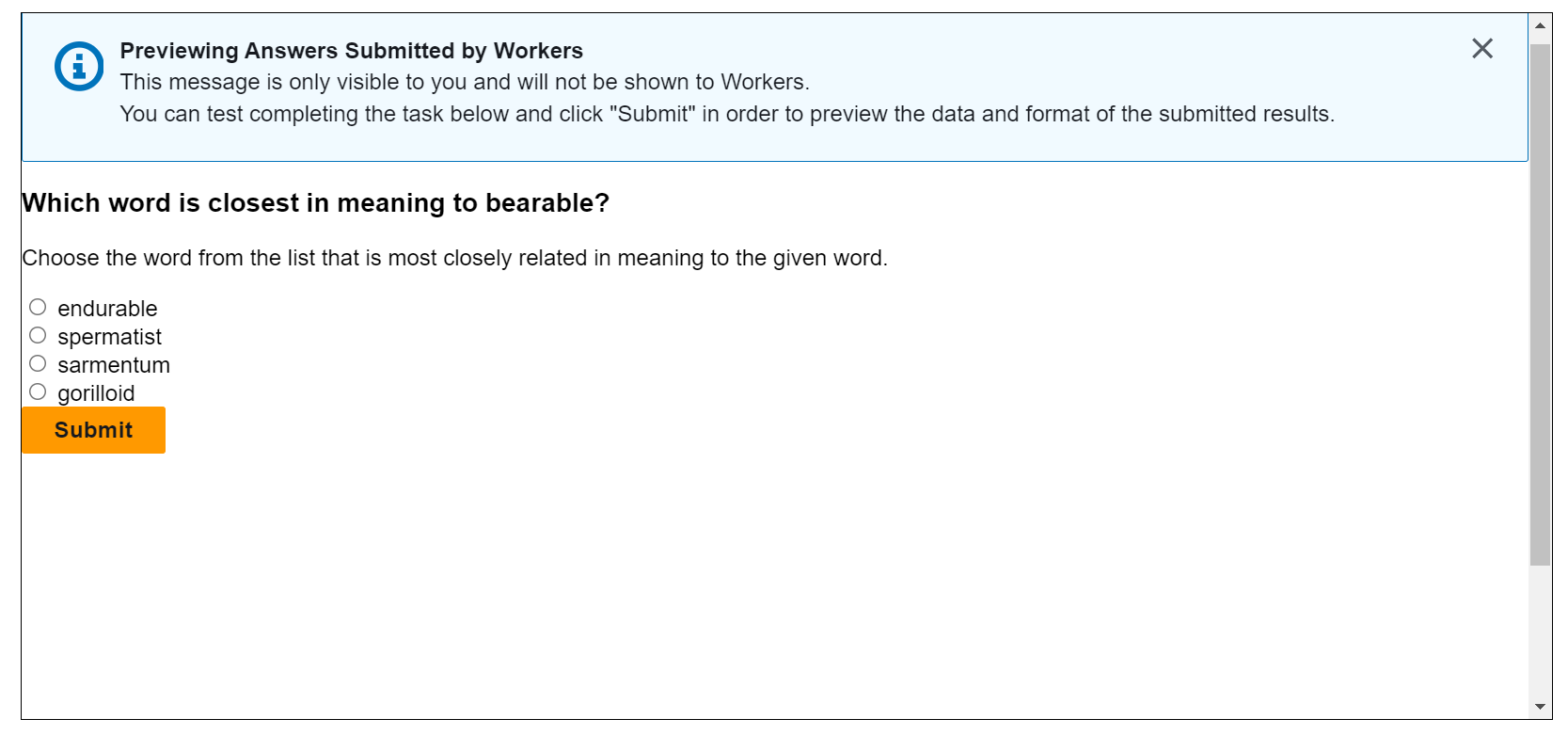}
  \includegraphics[scale=0.21]{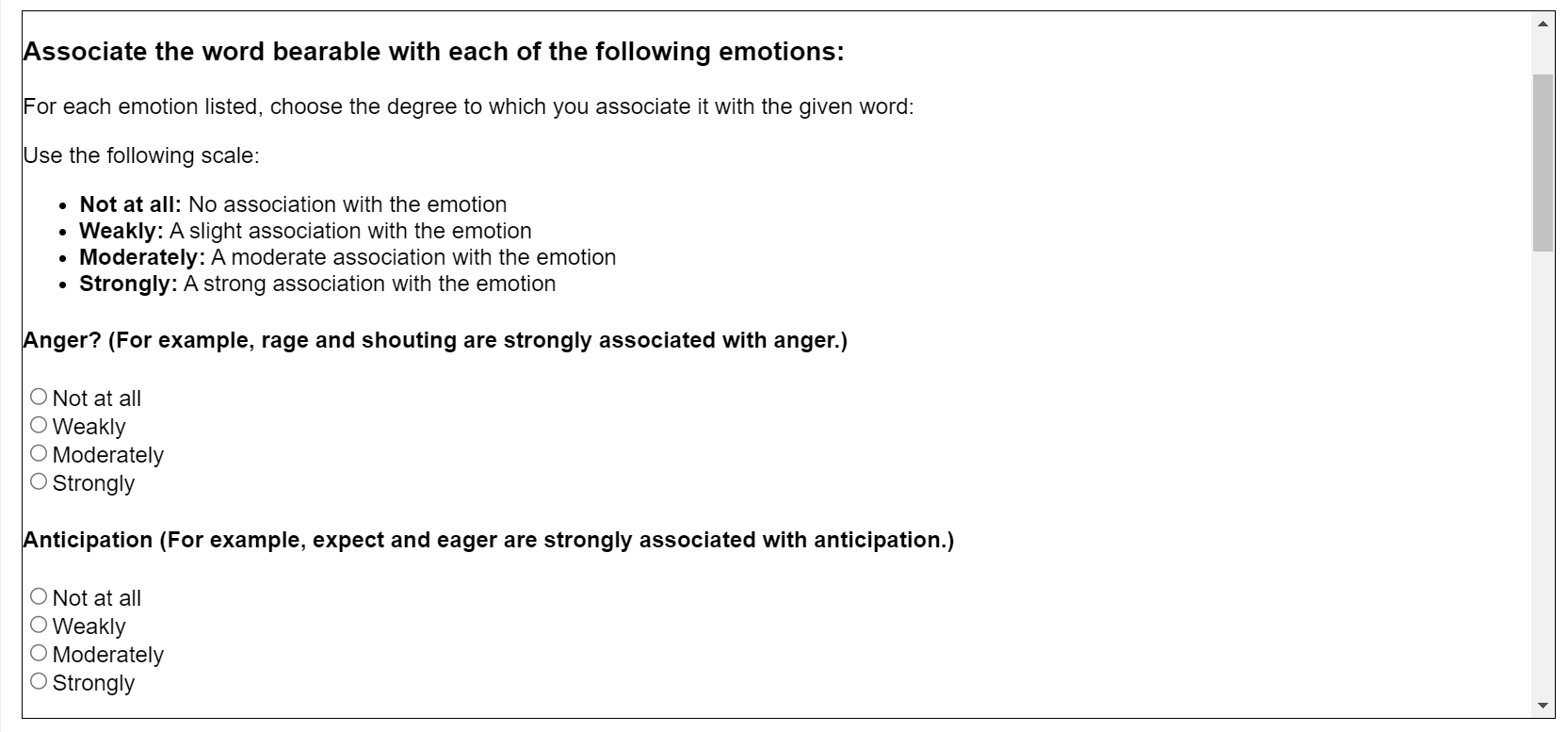}
  \includegraphics[scale=0.21]{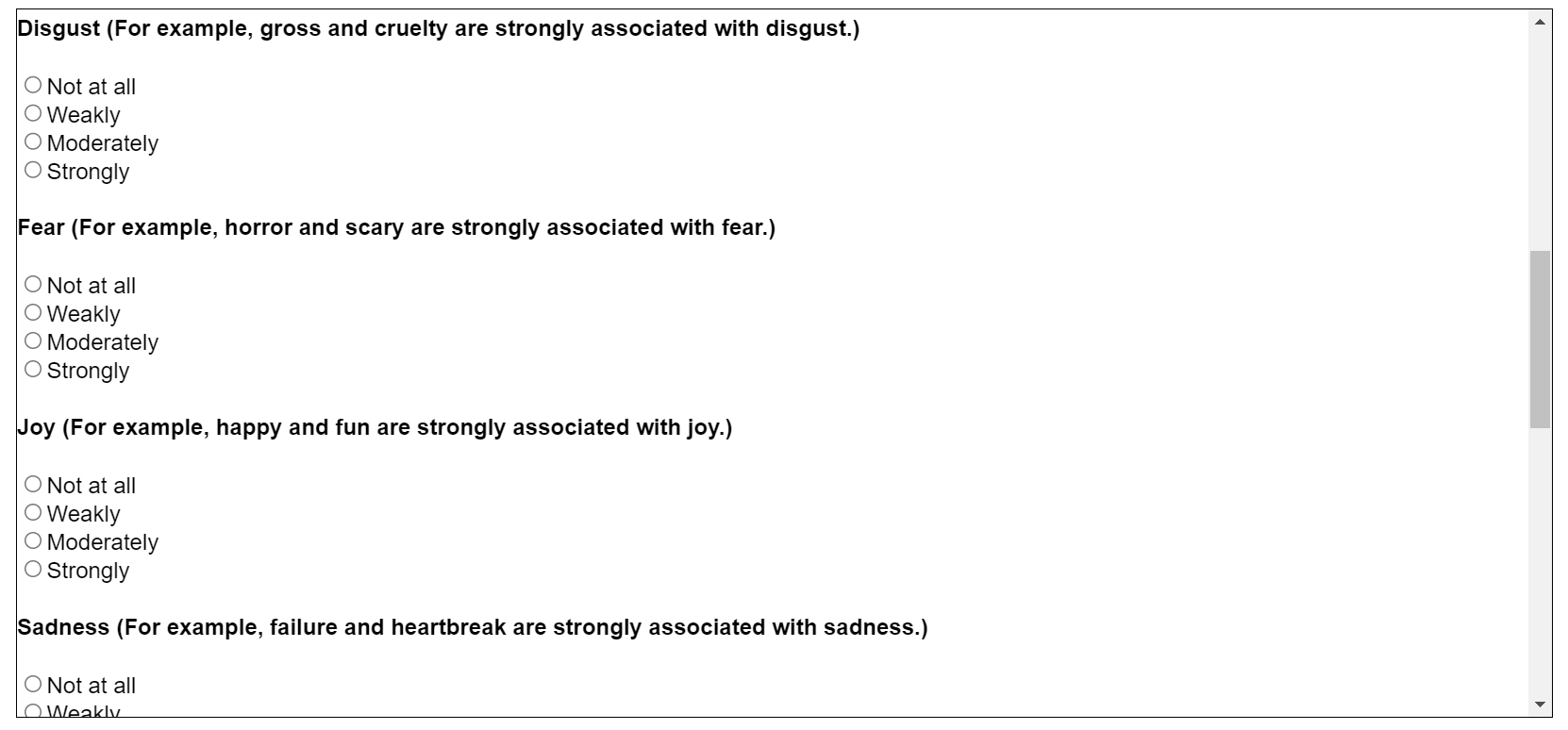}
  \includegraphics[scale=0.21]{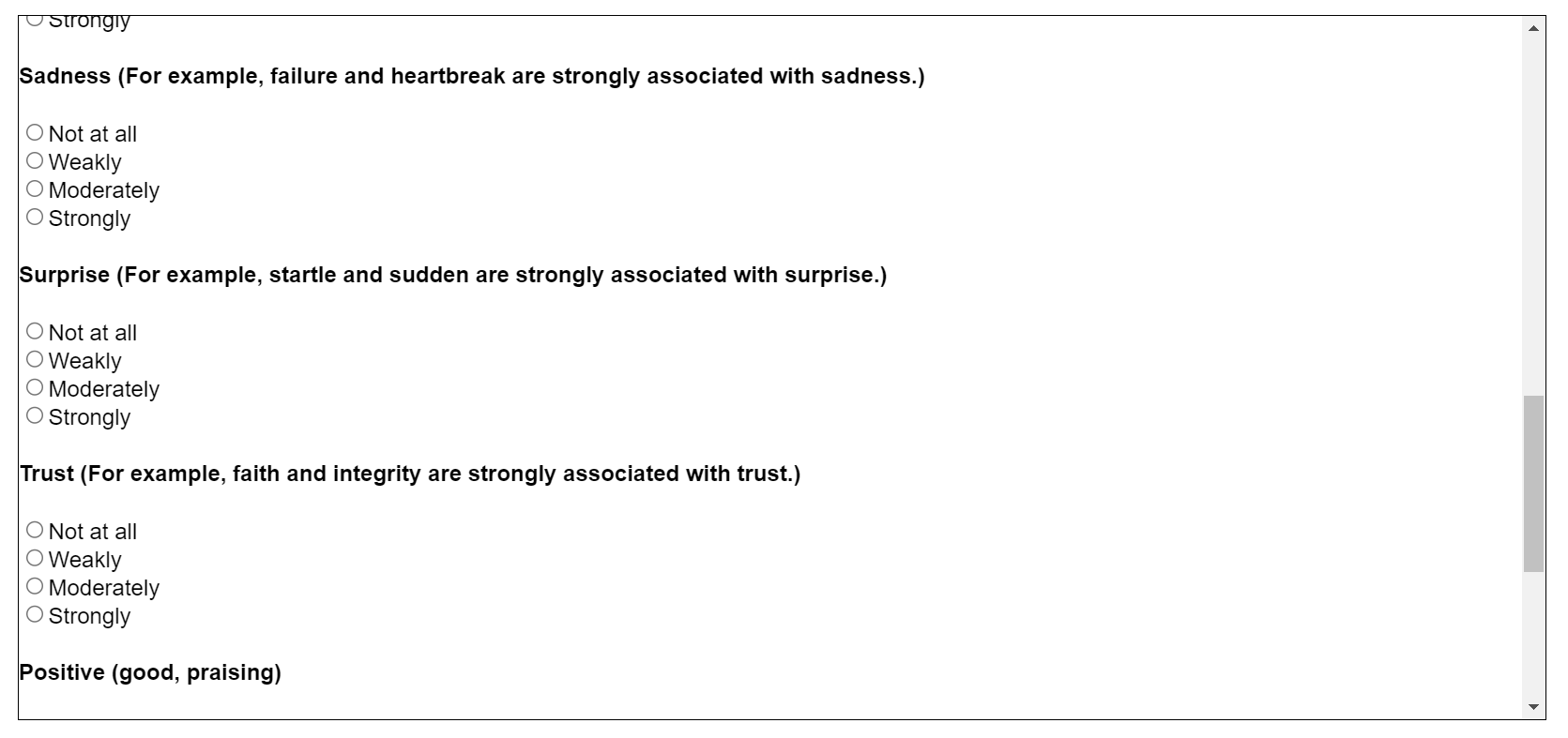}
  \includegraphics[scale=0.21]{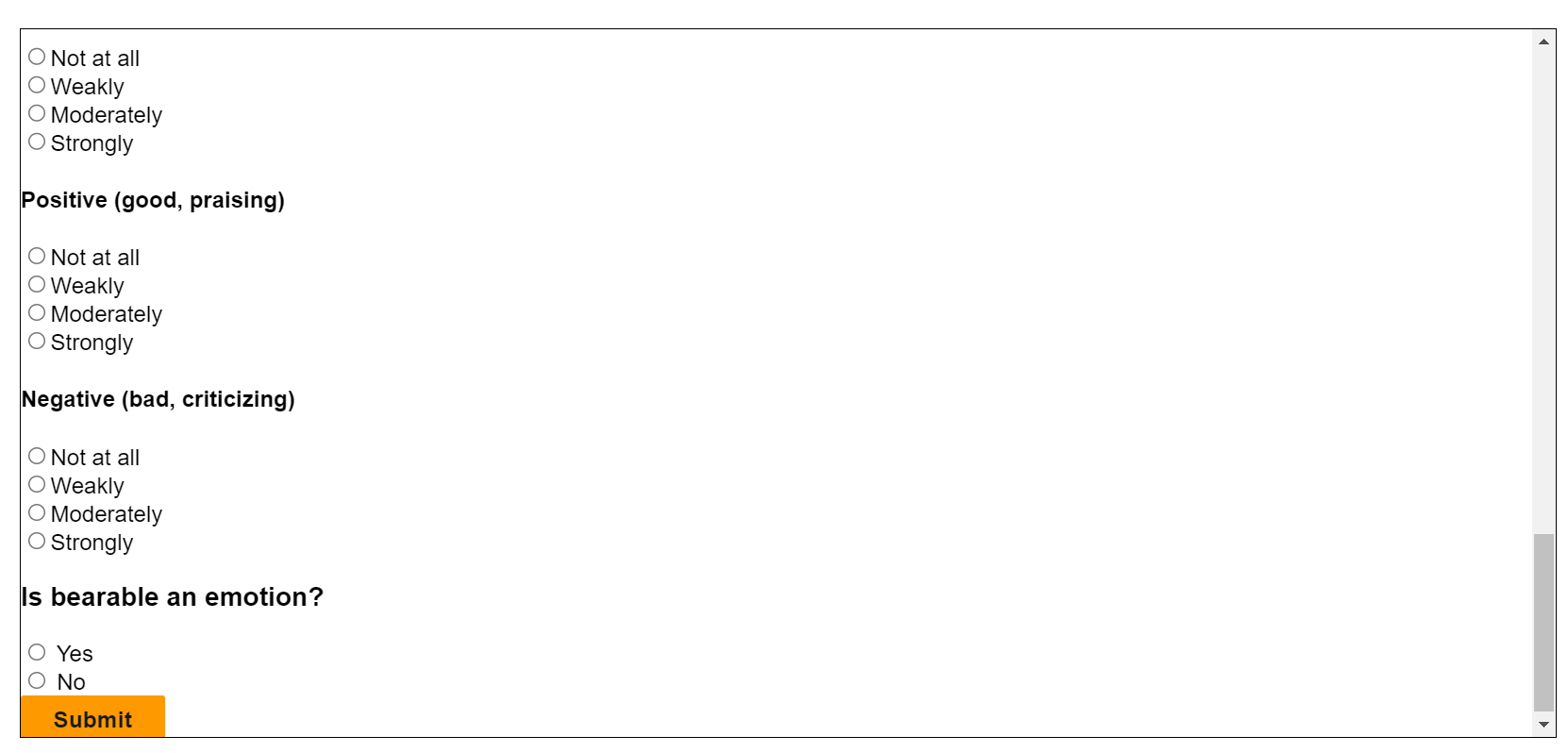}

  \caption{Amazon Mechanical Turk user interface for EmoLex annotations}
  \label{fig:turk-one}
\end{figure*}

Figures~\ref{fig:turk-one} include screenshots of the user interface created for workers to provide annotations for the 10 dimension emotion vector.

\begin{figure*}[!htb]
  \centering
  \includegraphics[scale=0.55]{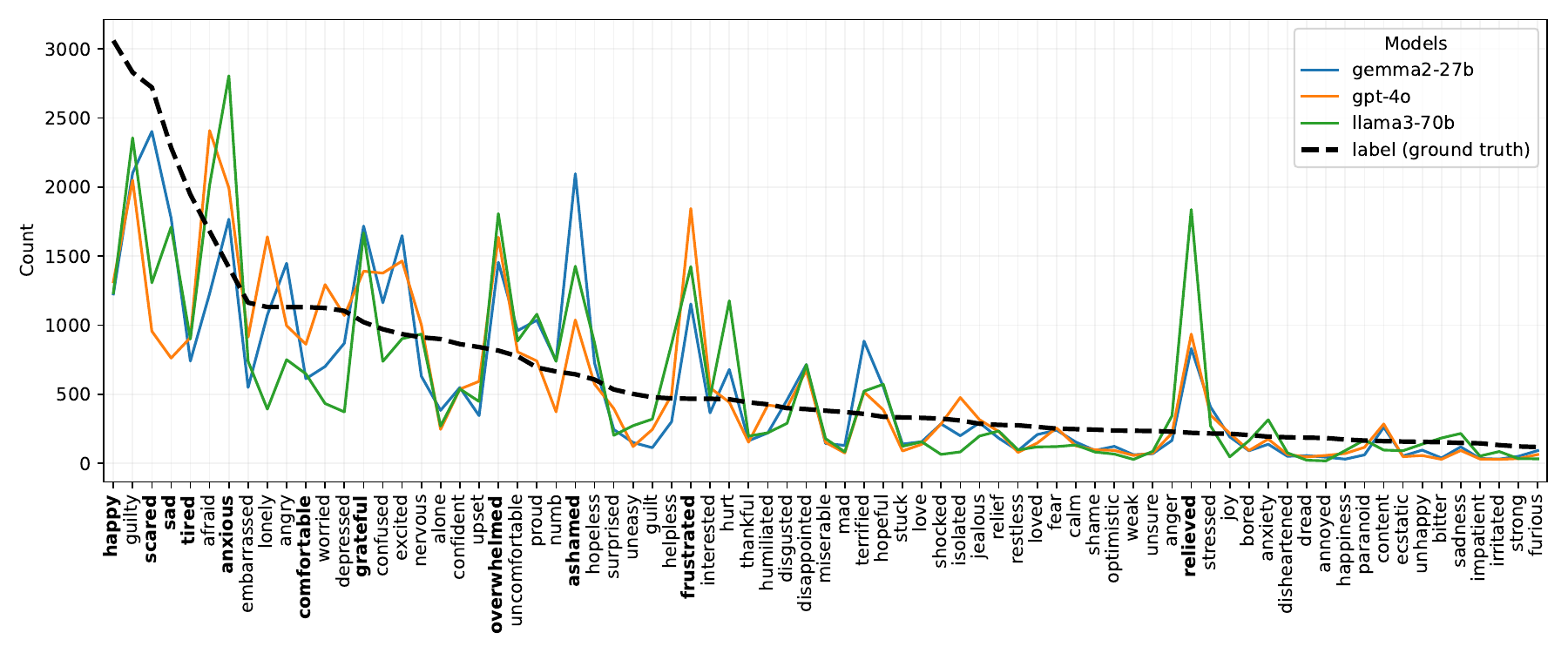}
\caption{Top 75 emotion counts: human labels vs. model predictions (GPT-4o, Llama 3.1-70B Instruct, Gemma 2-27B). Emotions are sorted by label frequency; larger differences are highlighted in bold.}
  \label{fig:label_pred_count}
\end{figure*}

\subsection{Statistical Modeling}

We conducted a Wilcoxon Signed-Rank Test to compare the LLMs' performance across the zero-shot, CoT, and two few-shot settings. This statistical test was applied to determine whether LLMs in the few-shot setting performed significantly better than those in the zero-shot setting, as well as whether the CoT setting performed significantly worse than the zero-shot setting. The LLMs, along with their corresponding p-values and Cohen’s d, are presented in Table~\ref{tab:statistical_modeling}.

In most cases, the few-shot settings outperform the zero-shot setting, while CoT settings perform worse than zero-shot. Only two settings do not show a significant difference from zero-shot: Llama-3.1-8B-instruct under the CoT setting and Llama-3.1-70B-instruct under the few-shot (random) setting. Additionally, the few-shot (nearest) setting consistently yields higher Cohen’s d scores across all conditions.

To compare of performance of different models under zero-shot setting, we also conducted Wilcoxon Signed-Rank Tests across models. The results are presented in Table~\ref{tab:statistical_modeling_2}.

\begin{table*}[!htb]
\fontsize{7}{7}\selectfont
\scalebox{1}{
\begin{tabular}{l|l|l|l}
\multicolumn{1}{l|}{\textbf{Model Name}} & \textbf{Setting} & \textbf{p-value} & \textbf{cohen's D}    \\ \hline

\multirow{3}{*}{Llama-3.1-8B-instruct} 
  & Few-shot (random) & 7.08e-11 & 0.027\\  
  & Few-shot (nearest)& 7.32e-66 & 0.077\\  
  & CoT               & 1.0      & -0.013\\ 
\hline
\multirow{3}{*}{Llama-3.1-70B-instruct} 
  & Few-shot (random) & 1.0      & -0.34\\  
  & Few-shot (nearest)& 2.70e-06 & 0.015\\  
  & CoT               & 0.0      & 0.161\\ 
\hline
\multirow{3}{*}{Gemma-2-2B-it} 
  & Few-shot (random) & 1.39e-08 & 0.026\\  
  & Few-shot (nearest)& 2.12e-68 & 0.086\\ 
  & CoT               & 1.86e-246& 0.120\\ 
\hline
\multirow{3}{*}{Gemma-2-9B-it} 
  & Few-shot (random) & 5.91e-54 & 0.061\\  
  & Few-shot (nearest)& 1.82e-89 & 0.082\\ 
  & CoT               & 0.0      & 0.204\\ 
\hline
\multirow{3}{*}{Gemma-2-27B-it} 
  & Few-shot (random) & 0.000125 & 0.014\\  
  & Few-shot (nearest)& 1.79e-13 & 0.028\\ 
  & CoT               & 0.0      & 0.314\\ 
\hline
\multirow{3}{*}{GPT-3.5-turbo} 
  & Few-shot (random) & 7.99e-22 & 0.035\\  
  & Few-shot (nearest)& 1.11e-37 & 0.048\\ 
  & CoT               & 4.79e-234& 0.116\\ 
\hline
\multirow{3}{*}{GPT-4o} 
  & Few-shot (random) & 1.66e-53 & 0.058\\  
  & Few-shot (nearest)& 6.25e-111& 0.086\\ 
  & CoT               & 4.79e-234& 0.116\\ 
  
\end{tabular}
}
\caption{P-values of different models and settings.}
\label{tab:statistical_modeling}
\end{table*}

\begin{table*}[!htb]
\fontsize{7}{7}\selectfont
\scalebox{1}{
\begin{tabular}{l|l|l|l}
\multicolumn{1}{l|}{\textbf{Model Name 1}} & \textbf{Model Name 2} & \textbf{p-value} & \textbf{cohen's D}    \\ \hline

\multirow{2}{*}{Flan-t5-xxl} 
  & Llama-3.1-8B-instruct & 2.33e-34 & 0.084\\  
  & Gemma-2-9B-it & 2.32e-86 & 0.144\\  
\hline

\multirow{2}{*}{GPT-3.5-turbo} 
  & Llama-3.1-70B-instruct & 6.17e-22    & 0.043\\  
  & Gemma-2-27B-it& 1.37e-82 & 0.087\\  
\hline
\multirow{2}{*}{Gemma-2-27B-it} 
  & Gemma-2-2B-it & 0.0    & 0.468\\  
  & Gemma-2-9B-it& 2.31e-129 & 0.099\\ 
\hline
\multirow{1}{*}{Gemma-2-9B-it} 
  & Gemma-2-2B-it & 0.0    & 0.367\\  
\hline
\multirow{1}{*}{Llama-3.1-70B-instruct} 
  & Llama-3.1-8B-instruct & 0.0    & 0.233\\  
\hline
\multirow{1}{*}{GPT-4o} 
  & GPT-3.5-turbo & 2.173e-277    & 0.160\\  
  
\end{tabular}
}
\caption{P-values of different models and settings.}
\label{tab:statistical_modeling_2}
\end{table*}

\end{document}